\pdfoutput=1

\documentclass[11pt]{article}

\usepackage[final]{acl}

\usepackage{times}
\usepackage{latexsym}

\usepackage[T1]{fontenc}

\usepackage[utf8]{inputenc}

\usepackage{microtype}

\usepackage{inconsolata}

\usepackage{graphicx}

\usepackage{amsmath,amsfonts,bm}

\def\eqref#1{equation~\ref{#1}}

\def\1{\bm{1}}

\DeclareMathAlphabet{\mathsfit}{\encodingdefault}{\sfdefault}{m}{sl}
\SetMathAlphabet{\mathsfit}{bold}{\encodingdefault}{\sfdefault}{bx}{n}

\usepackage{graphicx}
\usepackage{hyperref}
\usepackage{url}
\usepackage{booktabs}
\usepackage{natbib}
\usepackage{amsthm}
\usepackage{algorithm}
\usepackage{algpseudocode}
\usepackage{multirow}
\usepackage{svg}
\usepackage{enumerate}
\newtheorem{lemma}{Lemma}
\newtheorem{proposition}{Proposition}
\usepackage{caption}
\usepackage{subcaption}
\usepackage{comment}
\usepackage{amsmath}

\theoremstyle{definition}
\newtheorem{definition}{Definition}[section]

\title{\centering Hallucination Detox: \\ Sensitivity Dropout (SenD) for \\ Large Language Model Training}

\author{
  \textbf{Shahrad Mohammadzadeh\textsuperscript{1, 4, \thanks{Equal Contribution}}},
  \textbf{Juan David Guerra\textsuperscript{2, 4, \textsuperscript{$ \ast $}}},\\
  \textbf{Marco Bonizzato\textsuperscript{2, 3, 4}},
  \textbf{Reihaneh Rabbany\textsuperscript{1, 4, 5}},
  \textbf{Golnoosh Farnadi\textsuperscript{1, 3, 4, 5}}\vspace{1mm}\\
  \textsuperscript{1}McGill University,
  \textsuperscript{2}Polytechnique Montréal,
  \textsuperscript{3}Université de Montréal,\\
  \textsuperscript{4}Mila - Quebec Artificial Intelligence Institute
  \textsuperscript{5}CIFAR AI Chair\vspace{1mm}\\
  \texttt{shahrad.mohammadzadeh@mail.mcgill.ca, juan-2.guerra@polymtl.ca}
  }

\begin{document}

\maketitle

\begin{abstract}
As large language models (LLMs) become increasingly prevalent, concerns about their reliability, particularly due to hallucinations - factually inaccurate or irrelevant outputs - have grown. Our research investigates the relationship between the uncertainty in training dynamics and the emergence of hallucinations. Using models from the Pythia suite and several hallucination detection metrics, we analyze hallucination trends and identify significant variance during training. To address this, we propose \textbf{Sensitivity Dropout (SenD)}, a novel training protocol designed to reduce hallucination variance during training by deterministically dropping embedding indices with significant variability. In addition, we develop an unsupervised hallucination detection metric, Efficient EigenScore (EES), which approximates the traditional EigenScore in 2x speed. This metric is integrated into our training protocol, allowing SenD to be both computationally scalable and effective at reducing hallucination variance. SenD improves test-time reliability of Pythia and Meta's Llama models by up to 17\% and enhances factual accuracy in Wikipedia, Medical, Legal, and Coding domains without affecting downstream task performance. 
\end{abstract}

\section{Introduction}

\subsection{Motivation}

As Large Language Models (LLMs) become more sophisticated and widespread across industries, concerns about their reliability and safety have grown due to misuse and user errors. One of these concerning areas discovered by the scientific community is the phenomenon of hallucinations - LLMs producing content that may not align with real-world facts, the user’s input, or training data it has seen in the past \citep{huang_survey_2023}. In our research we target confabulations, hallucinations which occur when the LLM generates different responses given the same or similar inputs. This can be harmful when the generations alter between correct and factually incorrect responses.

Previous research has largely focused on identifying and addressing hallucinations in large language models (LLMs), but the impact of the training process on hallucinations remains under-explored \citep{huang_survey_2023, rawte_survey_2023, ye_cognitive_2023, hong_hallucinations_2024, xu_hallucination_2024, chen_inside_2024, li_dawn_2024, gao_retrieval-augmented_2024}. This paper investigates how iterative learning in LLMs causes significant variance in hallucination behavior, leading to fluctuating prediction confidence and making it difficult to identify a checkpoint where the model reliably learns facts.


To explore these hallucination trends, we analyze models ranging from 70 million to 12 billion parameters within the Pythia suite \citep{biderman_pythia_2023}, assessing them across various training checkpoints and tasks. Our goal is to validate the oscillatory behavior observed by \citet{li_dawn_2024} through evaluation metrics including HaluEval \citep{li_halueval_2023}, FactScore \citep{min_factscore_2023}, SelfCheckGPT \citep{manakul_selfcheckgpt_2023}, and XSum \citep{narayan_dont_2018}. Utilizing the reliability of internal model dynamics for quantifying hallucination likelihood, we use EigenScore \citep{chen_inside_2024} and Semantic Entropy \citep{kossen_semantic_2024} to detect hallucination risk by analyzing variability in high-temperature outputs. Experiments utilize EigenScore and the HELM dataset \citep{su_unsupervised_2024} to identify hallucinations during training. 


We introduce \textbf{Sensitivity Dropout} (SenD), a novel training protocol that prioritizes confident learning over mere loss minimization. SenD reduces hallucination variance by selectively dropping Sensitive Embedding Indices,—those exhibiting significant fluctuations throughout training— improving model certainty and providing a reliable stopping criterion for training. To enhance efficiency, we propose \textbf{Efficient EigenScore} (EES), a scalable alternative to EigenScore \citep{chen_inside_2024} for hallucination detection, maintaining high correlation while reducing computational costs.

Our contributions to the field, emphasizing that SenD enhances the training process and \textbf{does not replace} post-hoc solutions, \footnote{For the code and datasets used, refer to our GitHub repository at: \url{https://github.com/EMZEDI/SEND}.} can be summarized as follows: 
\begin{enumerate}
    \item Empirical verification of the \textbf{hallucinatory oscillation in LLM training} across various model scales and detection metrics.
    \item  \textbf{Sensitivity Dropout  (SenD)}, a novel training paradigm designed to reduce hallucination variance and increase model confidence during training.
    \item \textbf{Efficient EigenScore (EES)}, an efficient hallucination detection metric used to keep SenD efficient, achieving up to 2x speedup with minimal effects on accuracy.
\end{enumerate}

\subsection{Related Work}
The majority of research on hallucinations in language models has focused on detecting and mitigating this phenomenon rather than explaining its underlying causes. Recent techniques can be categorized into two main approaches: those based on output probabilities at inference time \citep{manakul_selfcheckgpt_2023, joshi_triviaqa_2017, li_halueval_2023} and those that utilize internal representations or hidden layers of the model \citep{su_unsupervised_2024, chen_inside_2024, kossen_semantic_2024}. While the former has shown effectiveness, the latter offers deeper insights, but often comes with computational trade-offs. Additionally, methods like Reinforcement Learning with Human Feedback (RLHF) have gained traction for enhancing model reliability \citep{yu_rlhf-v_2024}. However, many of these post-hoc solutions enhance factual accuracy by layering algorithms atop pre-trained models, which can be inefficient. Our work addresses this gap by focusing on the internal dynamics of the model that contribute to hallucinations.

Regularization techniques have been introduced to fix the issue of variability, notably random neuron dropout, used to reduce the variance and ensure that no neuron is overpowering others \citep{srivastava_dropout_2014, baldi_understanding_2013}. Work such as that done by \citet{santra_deterministic_2020, ba_adaptive_2013} aims to modify random neuron dropout to change the way neurons are dropped to a more deterministic, precise manner. This has allowed the authors to drop unimportant connections in a deep neural network to ensure that class discriminative information is propagated through the model correctly \citep{santra_deterministic_2020}. Inspired by this, our aim is to target hallucinatory embedding indices in our models to ensure that information is learnt with certainty. State-of-the-art hallucination metrics, especially those based on internal model dynamics, rely on spectral analysis and embedding matrix computations. Methods like EigenScore \citep{chen_inside_2024} and Semantic Entropy \citep{kossen_semantic_2024} effectively assess hallucination risk but require multiple inferences, making them computationally demanding as models scale. Tools such as the Density of States (DOS) and the kernel polynomial method (KPM) have been explored to approximate spectral properties efficiently \citep{huang_fast_2023, lin_approximating_2014}. Building on these advancements, our work integrates efficient spectral analysis methods into hallucination detection, demonstrated through EES and SenD.

\section{Oscillatory Behaviour Validation}

\begin{figure}[ht]
\centering
\begin{subfigure}[b]{0.494\columnwidth}
\centering
  \includegraphics[width=\linewidth]{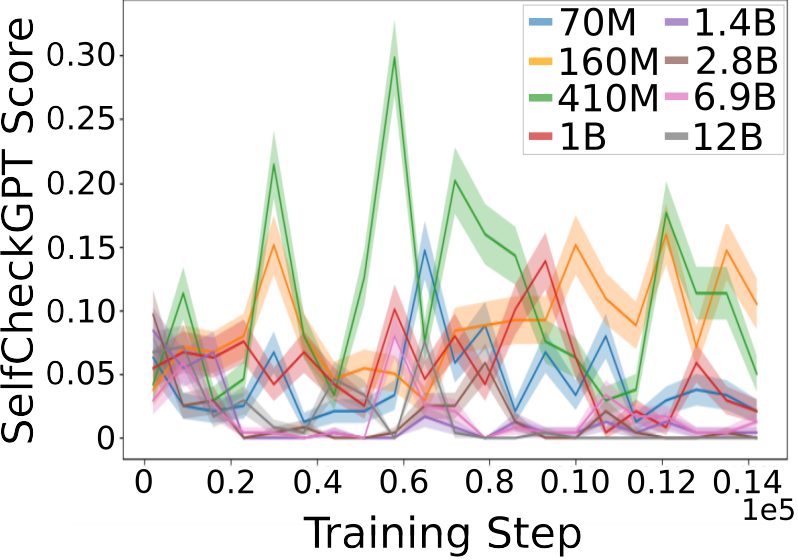}
    \caption{Self-Consistency}
  \label{fig:dynamics_oscillations}
\end{subfigure}
\hfill
\begin{subfigure}[b]{0.494\columnwidth}
  \includegraphics[width=\columnwidth]{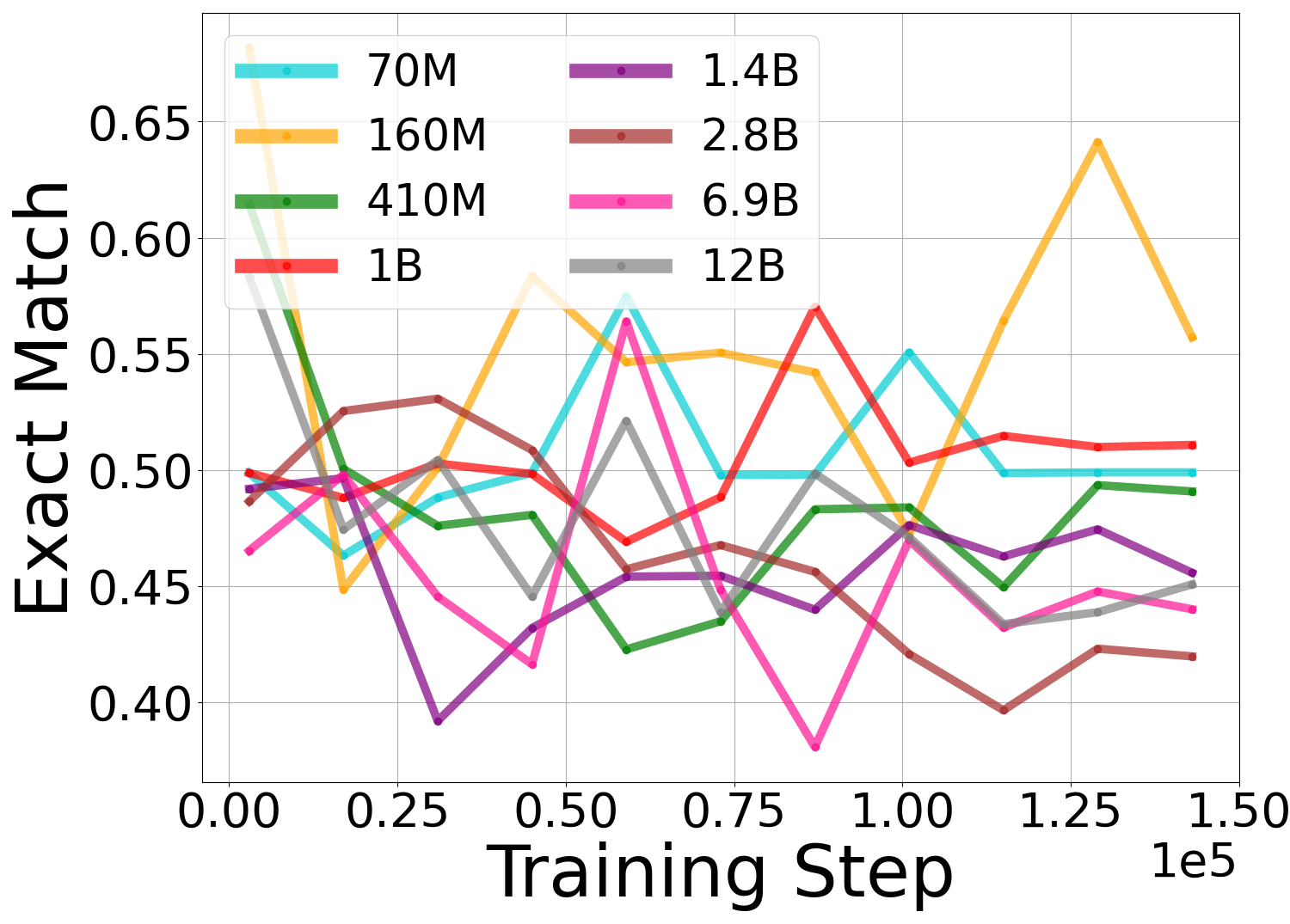}
  \caption{QA}
  \label{fig:dynamics_models}
\end{subfigure}
\caption{\textbf{Visualization of Oscillatory Behavior}. (a) SelfCheckGPT and (b) HaluEval EM metrics across model sizes from 70M-12B. Solid lines show average performance, shaded regions indicate standard deviation. High variance in hallucination metrics highlights the need for stabilization. For Perplexity (PPL), Rouge1 and other HaluEval metrics refer to Appendix \ref{appendix:oscillations}.}
\label{fig:dynamics}
\end{figure}

Transformer training checkpoints are vital for understanding learning dynamics. Our analysis shows that converged training loss does not necessarily reduce hallucinations, confirming \citet{li_dawn_2024}'s observations on LLM oscillatory hallucination behavior. We leverage Eleuther AI’s Pythia and LMEval tools \citep{biderman_pythia_2023, gao_framework_2024} to study 16 LLMs (ranging from 70M to 12B parameters) across 20 evenly spaced checkpoints. At each checkpoint, we evaluate the models on various hallucination metrics: SelfCheckGPT for self-consistency \citep{manakul_selfcheckgpt_2023}, XSum for summarization \citep{narayan_dont_2018}, perplexity, and HaluEval for QA tasks \citep{li_halueval_2023}. As state-of-the-art (SOTA) methods, we select to present SelfCheckGPT and HaluEval QA Exact Match in Figure \ref{fig:dynamics} as representatives of the set of introduced metrics as they exhibit similar behaviours to the rest of the metrics in Appendix \ref{appendix:oscillations}. Higher SelfCheckGPT scores indicate more self-contradiction, higher Rouge1 on XSum suggests better summary alignment, lower perplexity implies greater prediction confidence, and improved QA scores reflect better overall performance.

\paragraph{RQ1: How do the established iterative training processes and model complexity influence LLM hallucinations?}
The analysis of hallucination oscillations, as shown in Figure \ref{fig:dynamics}, indicates a consistent pattern across different models: oscillations persist throughout training from the initial to the final checkpoint. This finding highlights the uncertainty of halting training solely based on the convergence of training loss. For instance, in QA settings, the optimal Exact Match of the outputs with ground truths is achieved in earlier checkpoints. This observation is seen more dramatically in Figure \ref{fig:dynamics_oscillations}, where the size of the model has almost no effect on the performance of SelfCheckGPT. Instead, we observe oscillatory behaviour within self-consistency, implying that model size is not much effective at tackling the issue of confabulations verified by results in Appendix \ref{appendix:oscillations} as well. These results suggest that optimizing solely for the loss in training is not sufficient. We also see that beyond a certain point, larger models do not significantly reduce hallucinations, indicating that scaling alone is not sufficient for building robust models. Instead, more refined approaches are needed to address the underlying variability in model behavior.

\section{Internal Training Dynamics}

Following our investigation of the oscillatory behaviour in training, we look into the internal states of the Pythia 1B model \citep{biderman_pythia_2023} to see what information we are able to extract. In doing so, we establish a series of definitions and metrics in order to understand the internal processes during the training of LLMs. This information is later used in Sections \ref{sec:ees} and \ref{sec:send} to assist us in deriving methods for improving the variance in the hallucinatory behaviour of models during training.

\subsection{Sensitive Embedding Indices}

To start our analysis of the internal states, we employ \citet{su_unsupervised_2024}'s sentence embedding extraction approach given its demonstrated success in hallucination detection. We convert the activation matrix of the model into a sentence embedding vector (Definition \ref{definition:sentence_embedding_vector}) which turns an \(\mathbb{R}^{n,m}\) activation matrix into a sentence embedding vector $a_k$ for input $k$ with dimension \(\mathbb{R}^{n}\).

\begin{definition}[Sentence Embedding Vector]
\label{definition:sentence_embedding_vector}
The Sentence Embedding Vector is a way to convert the large \(\mathbb{R}^{n,m}\) activation matrix into a smaller, easier to manage vector with dimension
    \(\mathbb{R}^{n}\).
    \begin{equation}
    e_k = {\frac{1}{2}}((\frac{1}{m} \sum_{i=1}^m H_{N-1}^i)+H_{N-1}^m)
    \end{equation}
Where $e_k$ is the activation of one input $k$, $m$ is the number of tokens in the sequence,  $H$ is the token embedding activation matrix, and $N-1$ is the subtraction to get the penultimate layer index and the formula is adapted from \citet{su_unsupervised_2024}. The penultimate layer of the LLM, being the layer closest to the output probabilities, is our primary focus for hallucination analysis due to its rich information about output certainty.
\end{definition}

Next, we define the Net Change Formula \ref{definition:net_change_formula} as a way to extract information from the model indicative of oscillatory behaviour between checkpoints from the sentence embedding vector.
\begin{definition}[Net Change Formula]
\label{definition:net_change_formula}
Let $e^t_i$ denote the embedding of data point $x$ at embedding index $i$ of the contextual embedding after checkpoint $t$. Then we define the net change formula as 
\begin{equation}
\Delta e_i^t=|e_i^t - e_i^{t-1}|
\end{equation}
\end{definition}

Building on these definitions, we now formalize the central focus of our investigation: \textbf{Sensitive Embedding Indices (SEIs)}, which we demonstrate play a critical role in the hallucination behavior of large language models (LLMs) (see Section~\ref{sec:sensitive_impact}). Specifically, SEIs can be leveraged to refine training procedures, reducing hallucination variability during training and improving overall confidence at inference time. Conceptually, SEIs correspond to indices within the sentence embedding (Definition~\ref{definition:sentence_embedding_vector}) that exhibit significant fluctuations across training checkpoints, a phenomenon we hypothesize to be closely linked to the oscillatory nature of hallucination performance. Identifying the most sensitive embedding indices involves selecting the top $K\%$ of indices for a given data point’s representation. In our study, we set $K = 20$.  

\begin{definition}[Sensitive Embedding Indices - SEIs]
\label{definition:sensitive_neurons}
Indices of the contextual embedding for data point $x$ which exhibit the highest net change across the last $C$ checkpoints of training, indicating overall high variability during this period. This is calculated by 
\begin{equation}
V_i=Var(e_i) \sum_{t=T-C+1}^T \Delta e_i^t
\label{equation:3}
\end{equation}
where $V_i$ is the total variability during the last $C$ checkpoints and the most sensitive embedding indices are
\begin{equation}
    \textbf{s} = \arg\max_{1 \leq i \leq N} \left\{ V_i \mid V_i \geq \text{percentile}(V, 100 - k) \right\}
\label{equation:4}
\end{equation}
where N is the embedding vector size and $k$ is the desired percentile threshold.
\end{definition}

The aforementioned definition of Sensitive Embedding Indices (SEIs) is subsequently applied to LLM hallucinations through an analysis of EigenScores. \citet{chen_inside_2024} introduce a novel metric for detecting confabulations, a specific subclass of hallucinations. Their approach computes an EigenScore (Definition~\ref{defintion:EigenScore}) by leveraging determinant calculations derived from multiple LLM outputs generated under a high-temperature setting (\textit{temperature} = 0.5), thereby encouraging greater output diversity. They hypothesize that when an LLM hallucinates, the resulting text exhibits increased semantic variability, leading to an elevated EigenScore. Notably, this method achieves SOTA performance while remaining unsupervised, as it relies solely on the model’s learned representations. In the following sections, we examine the correlation between EigenScores at various training checkpoints and the most sensitive embedding indices associated with the corresponding data points.

\begin{definition}[EigenScore]
\label{defintion:EigenScore}
    The \textbf{EigenScore} of data point $x$ indicates the degree of hallucination on input \( x \) by the average logarithm of the eigenvalues on the covariance matrix of the multiple output generations (typically 10 in our experiments).
    \begin{equation}
    \label{equation:5}
    ES =
    \mathbb{E}(Y \mid x, \theta) = \frac{1}{K} \sum_{i=1}^K \log(\lambda_i)
    \end{equation}
    where \( \lambda = \{ \lambda_1, \ldots, \lambda_K \} \) denotes the eigenvalues of the regularized covariance matrix \( \Sigma + \alpha \cdot \mathbb{I} \). we advise referring to \citet{chen_inside_2024} for a more detailed analysis of this formula.
\end{definition}

\label{sec:sensitive_impact}
\paragraph{RQ2: What is the impact of Sensitive Embedding Index on EigenScores and hallucination in LLMs?}

To assess the correlation between SEIs and other indices in the embedding matrix of 10 generated outputs at a specific checkpoint, we conduct experiments aimed to determine if the presence of SEIs indicates higher uncertainty and a greater likelihood of hallucinations.

\begin{figure}[ht]
\centering
\begin{subfigure}[b]{0.494\columnwidth}
\centering
  \includegraphics[width=\columnwidth]{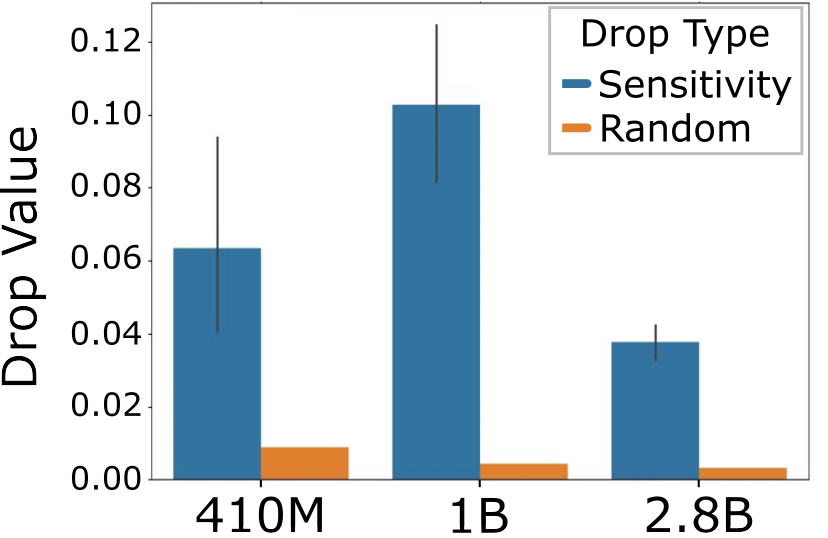}
    \caption{Model Size}
  \label{fig:drop_model_size}
\end{subfigure}
\hfill
\begin{subfigure}[b]{0.494\columnwidth}
\centering
  \includegraphics[width=\columnwidth]{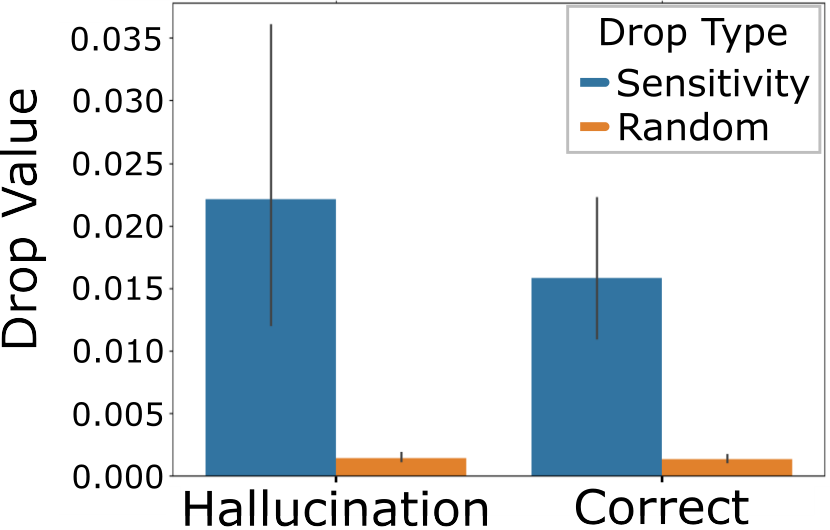}
    \caption{Output Type}
  \label{fig:drop_hallu_type}
\end{subfigure}
\caption{\textbf{Comparison of sensitive embedding index dropout} on inference of Eleuther AI's Pythia modeld with random embedding index dropout. The Y axis, Drop Value, denotes the decrease in EigenScore, ie less confabulations, following the dropout method used. (a) SEI dropout results are consistent across model sizes. (b) Hallucinatory outputs show a larger EigenScore drop than correct ones. SEI dropout significantly reduces EigenScore compared to random dropout in both (a) and (b)}
\label{fig:drop}
\end{figure}

To evaluate the effect of SEIs on hallucination, we conduct experiments using the HELM dataset \citep{su_unsupervised_2024}, which is comprised of model-generated outputs from over 50,000 Wikipedia articles. This dataset was selected due to Wikipedia's role as a primary factual source. 

To quantify the influence of SEIs, we extend the EigenScore method by applying it to sentence embeddings extracted from the penultimate layer of EleutherAI’s Pythia 1B model \citep{biderman_pythia_2023}. Our analysis focuses on checkpoints between 133,000 and 143,000 training steps, where embeddings exhibit greater stability and the model demonstrates a higher level of language understanding compared to earlier training phases. 

We implement SEI dropout by removing the top 10\% of SEIs at each checkpoint and compare it to a baseline where 10\% of embedding indices are randomly dropped. Furthermore, we examine the impact of SEI dropout on hallucination-prone versus non-hallucination-prone inputs to assess whether SEIs play a critical role in hallucination without adversely affecting factually correct outputs.

Given that a reduction in the EigenScore metric serves as a proxy for decreased hallucination likelihood, we adopt this metric as the primary evaluation measure in our study. Through a comparative analysis of baseline random embedding index dropout and SEI dropout, we demonstrate that SEI dropout significantly lowers the EigenScore across model sizes, thereby reducing confabulation probability (Figure~\ref{fig:drop_model_size}). Notably, while this reduction is most pronounced in hallucinatory outputs, we observe a decrease for correctly answered queries (Figure~\ref{fig:drop_hallu_type}), suggesting that our approach effectively modulates uncertainty without adversely impacting factual responses. Furthermore, our findings indicate that the internal states of the model play an important role in mitigating the generation of confabulated text across various model sizes.

\subsection{Efficient EigenScore Approximation}
\label{sec:ees}

\begin{algorithm}[H]
\caption{Efficient EigenScore Algorithm}
\label{alg:ees}
\small
\begin{algorithmic}[1]
    \Require Embedding matrix $E \in \mathbb{R}^{d_{\text{model}} \times K}$, number of Chebyshev terms $M$, number of stochastic trace estimation samples $N_z$
    \Ensure Approximated EigenScore $\textit{EES}$

    \State \textbf{Standardize and Scale the Embedding Matrix $E$:}
    \State $E_{\text{mean}} = \frac{1}{K} \sum_{i=1}^{K} E[:, i]$ 
    \State $E_{\text{std}} = \sqrt{\frac{1}{K} \sum_{i=1}^{K} (E[:, i] - E_{\text{mean}})^2}$ 
    \State $E_{\text{normalized}} = \frac{E - E_{\text{mean}}}{E_{\text{std}}}$ \Comment{Normalize $E$ with mean and standard deviation}
    \State $\sigma_{\text{max}} = \text{Power Method}(E_{\text{normalized}})$ \Comment{Compute largest singular value using the power method}
    \State $E_{\text{normalized}} \gets \frac{E_{\text{normalized}}}{\sigma_{\text{max}}}$ \Comment{Scale $E$ by $\sigma_{\text{max}}$}

    \State \textbf{Initialize:}
    \State $d_m = 0 \quad \forall m \in \{0, 1, \dots, M\}$ \Comment{Initialize $d_m$ coefficients}
    \State $c_m = 0 \quad \forall m \in \{0, 1, \dots, M\}$ \Comment{Initialize $c_m$ coefficients}

    \State \textbf{Compute DOS coefficients $d_m$:}
    \For{$m = 0$ to $M$}
        \State \textbf{Sample $z_j \sim \mathcal{N}(0, I)$} \Comment{Sample random vectors for stochastic trace estimation}
        \State \textbf{Compute Chebyshev polynomial using the recurrence relation}
    \EndFor

    \State \textbf{Compute Chebyshev coefficients $c_m$:}
    \For{$m = 0$ to $M$}
        \State $c_m \gets \int_{0}^{1} \log(\lambda) T_m^*(\lambda) \, d\lambda$ \Comment{Using Equation \ref{equation:ees_main}} and Gaussian Quadrature for approximation
    \EndFor

    \State \textbf{Compute EigenScore:}
    \State $\textit{EES} \gets \frac{1}{K} \sum_{m=0}^{M} d_m c_m$ \Comment{Approximate EigenScore using DOS coefficients}

    \State \Return $\textit{EES}$ \Comment{Return the approximated EigenScore}
\end{algorithmic}
\end{algorithm}
\normalsize


If \( n \) denotes the hidden layer size, and computing the EigenScore for a single inference requires an eigen-decomposition with complexity on the order of \( O(n^3) \). For \( T \) inferences, the overall computational cost scales as \( O(T \cdot n^3) \), which quickly becomes prohibitive as both \( n \) and \( T \) increase. To address the computational complexity of EigenScore calculations, particularly as LLM hidden layer sizes increase, we develop an approximation method. This approximation, detailed in Algorithm \ref{alg:ees}, leverages the properties of Spectral Density or Density of States (DOS) to estimate EigenScore without explicitly constructing the covariance matrix. While this approximation provides a general overview of EigenScore trends, it is important to note that the output scales differ: EigenScore ranges from $[0, \infty)$, whereas the approximation, referred to as \textbf{Efficient EigenScore (EES)}, outputs values between $[-1, 1]$. Since the spectrum of the matrix is altered to make EES computable and operates on its own scale, EES can be seen as a standalone metric for hallucination detection.
    
The computation of the Efficient EigenScore (EES) is based on two fundamental concepts: Chebyshev Polynomials and Density of States (DOS). A detailed introduction to these concepts is provided in Appendix sections \ref{appendix:chebyshev} and \ref{appendix:dos}. Below, we outline a brief sketch of the derivation of EES. Since \citet{chen_inside_2024} uses the covariance matrix of the embedding matrix of 10 sequences generated by the model in their methods, we represent it with $H$ and use it in our derivation.

\begin{lemma}
\label{lemma:mu_trace}
Let \( f = \log \). Then, for a covariance matrix \( H \) with eigenvalues \(\lambda_i\), we have
\begin{equation}
\text{trace}(\log(H)) = \sum_{i=1}^{N} \log(\lambda_i),
\end{equation}
where \( \lambda_i \) are the eigenvalues of \( H \). 
\end{lemma}

\begin{proposition}
\label{proposition:eigenscore}
Using the property of the density of states (DOS), we have:
\begin{equation}
\label{equation:prop}
\int \log(\lambda) \, \mu(\lambda) \, d\lambda = \log \left( \prod_{i=1}^{N} \lambda_i \right),
\end{equation}
which follows from Lemma \ref{lemma:mu_trace} since \(\sum_{i=1}^{N} \log(\lambda_i) = \log \left( \prod_{i=1}^{N} \lambda_i \right)\).
\end{proposition}

Note that from Proposition \ref{proposition:eigenscore}, the integral is equal to $N . EigenScore(H)$ or in our application, given $C$ the integral equals $K . EigenScore(C)$, $K$ being the number of model generations.

Our objective is to simplify the integral and approximate its value, avoiding the direct computation of the covariance matrix. This approach is intended to mitigate the computational complexity and associated costs of explicitly handling the covariance matrix. Further utilizing Chebyshev Polynomials, DOS, and KPM (as introduced in Appendix \ref{appendix:dos}), we can simplify the integral mentioned in Equation \ref{equation:prop} to $\sum_{m=0}^{M} d_m c_m$, where $d_m$ term in DOS is approximated using Stochastic Trace Estimation and $c_m$ m'th Chebyshev Polynomial coefficient. Appendices \ref{section:stochastic} and \ref{section:integral_calculation} provide the derivation of this equation. Note that the simplified integral is ultimately used to approximate the EigenScore of the matrix which is ultimately equivalent to $\frac{1}{K} \sum_{m=0}^{M} d_m c_m$.  Performance of EES approximation is closely correlated with that of the original EigenScore which can be seen in Figure \ref{fig:finetuning_EigenScore}.

\begin{figure}[ht]
\begin{center}
  \includegraphics[width=\columnwidth]{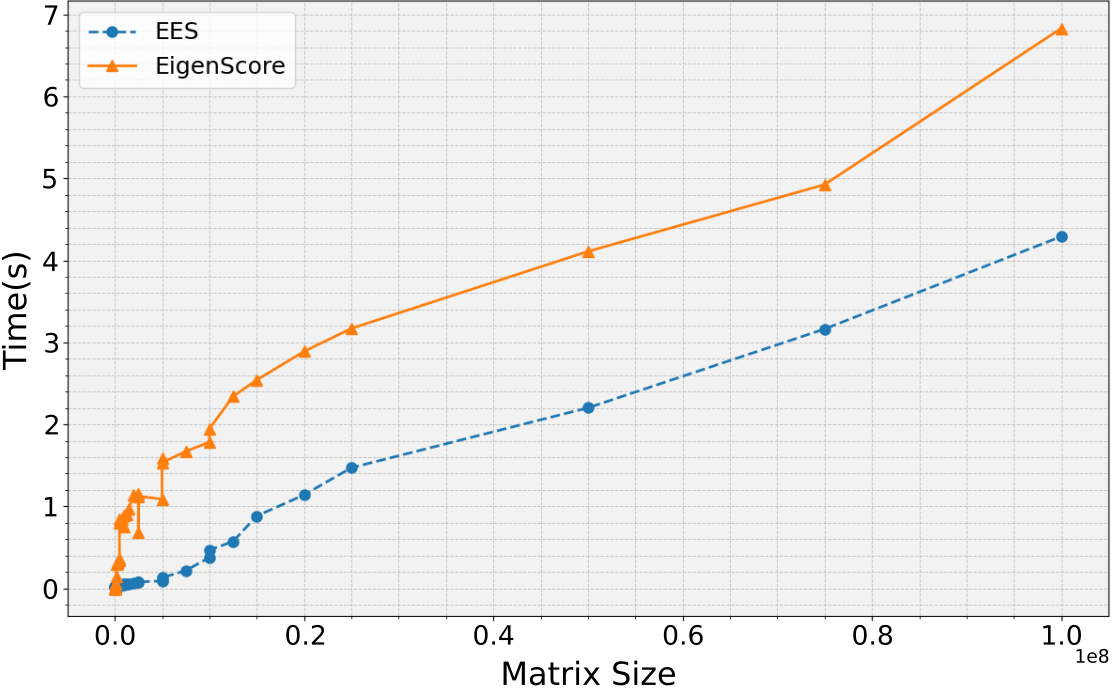}
\end{center}
\caption{\textbf{Efficient EigenScore approximation scaling}. Computation time comparison between EigenScore and EES (moments = 20). The x-axis represents matrix size (rows × columns), and the y-axis shows computation time. As matrix size increases, EES consistently reduces computation time, making it a practical choice.}
\label{fig:eesMatrix}
\end{figure}
\paragraph{RQ3: How does EES scale compared to regular EigenScore?}
The efficiency of EES is compared to that of the regular EigenScore calculation with respect to scaling matrix sizes. These tests are imperative to the application of our training protocol on   increasing LLM sizes  in Section \ref{sec:send}  due to  larger matrix sizes to decompose for the EigenScore calculation. We conduct a grid search over two important parameters: Matrix size (Figure \ref{fig:eesMatrix}) and Moments used for EES calculation (Figure \ref{fig:eesMoments}). The difference between EES time in comparison to EigenScore when increasing the number of columns and rows is visualized in Figure \ref{fig:eesMatrix} using a moments value of 20. It is evident that EES provides a significant computational advantage when increasing the number of columns or rows. Remarkably, at matrix size $\mathbb{R}^{1e8}$, EES nearly halves the computation time of regular EigenScore calculation at around 4 seconds whereas EigenScore takes approximately 7 seconds to calculate. We can then deduce that given a good enough approximation, EES provides a significant reduction in computational complexity as model and matrix size increase.

\begin{figure*}[t]
\centering
\begin{subfigure}[b]{0.24\textwidth}
\centering
  \includegraphics[width=\textwidth]{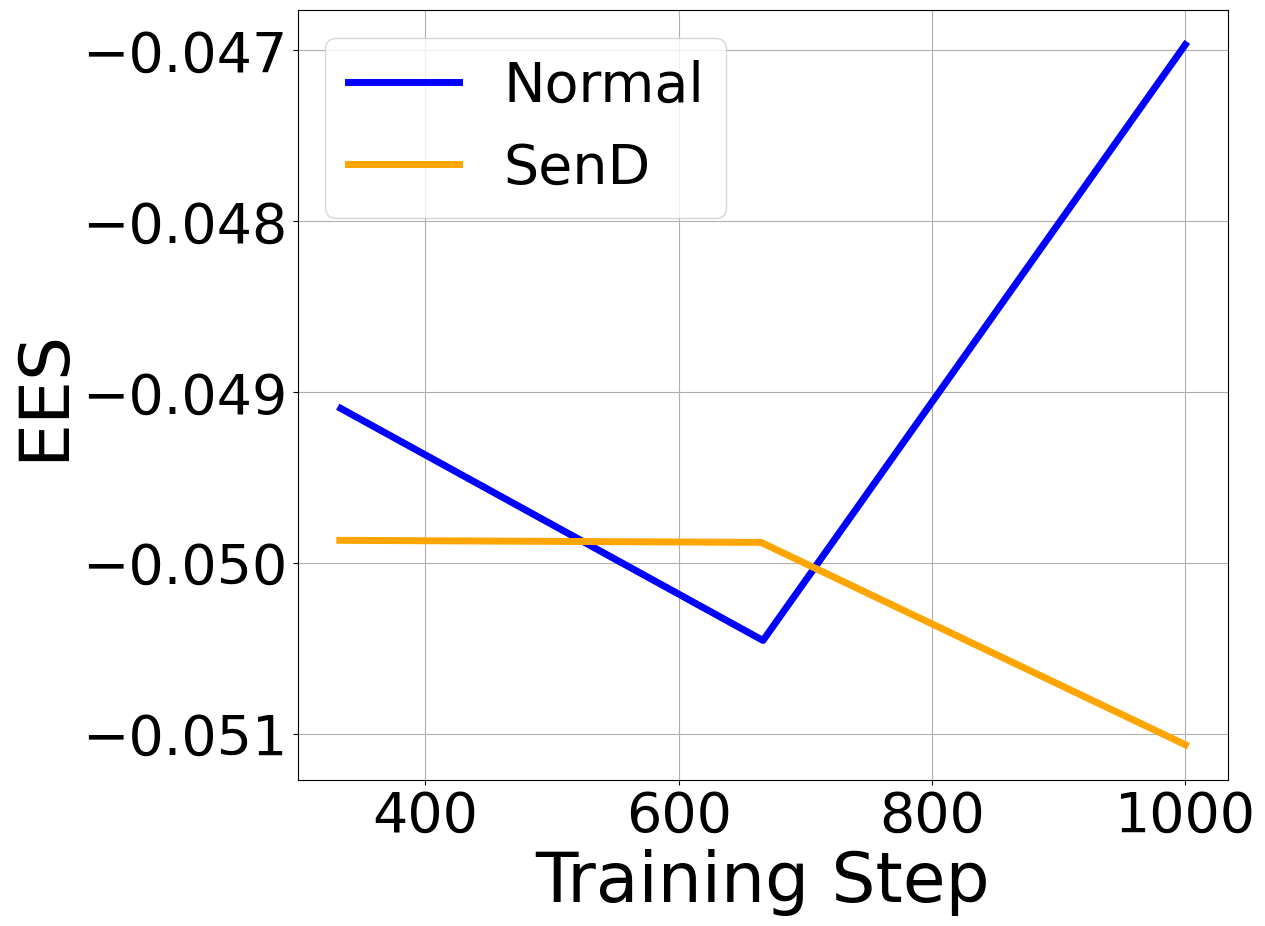}
  \caption{Llama 8B - HELM}
  \label{fig:llama_helm}
\end{subfigure}
\hfill
\begin{subfigure}[b]{0.25\textwidth}
\centering
   \includegraphics[width=\textwidth]{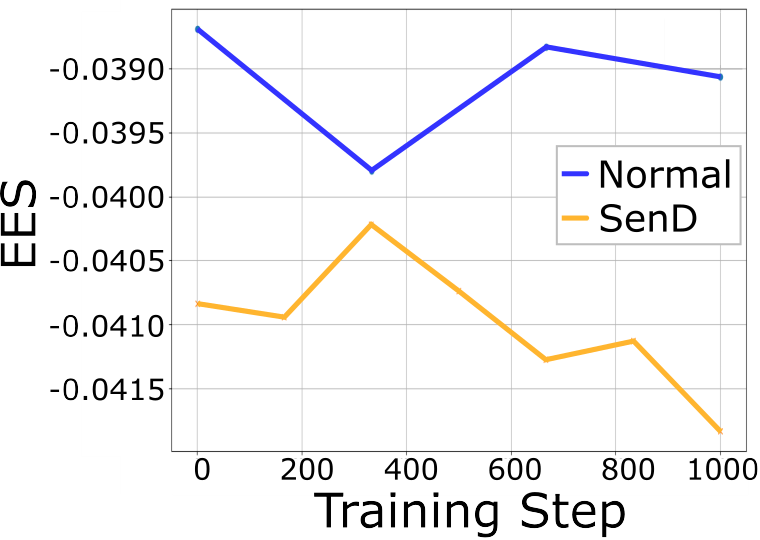}
  \caption{Llama 8B - LegalBench}
  \label{fig:llama_legalbench}
\end{subfigure}
\hfill
\begin{subfigure}[b]{0.24\textwidth}
    \centering
    \includegraphics[width=\textwidth]{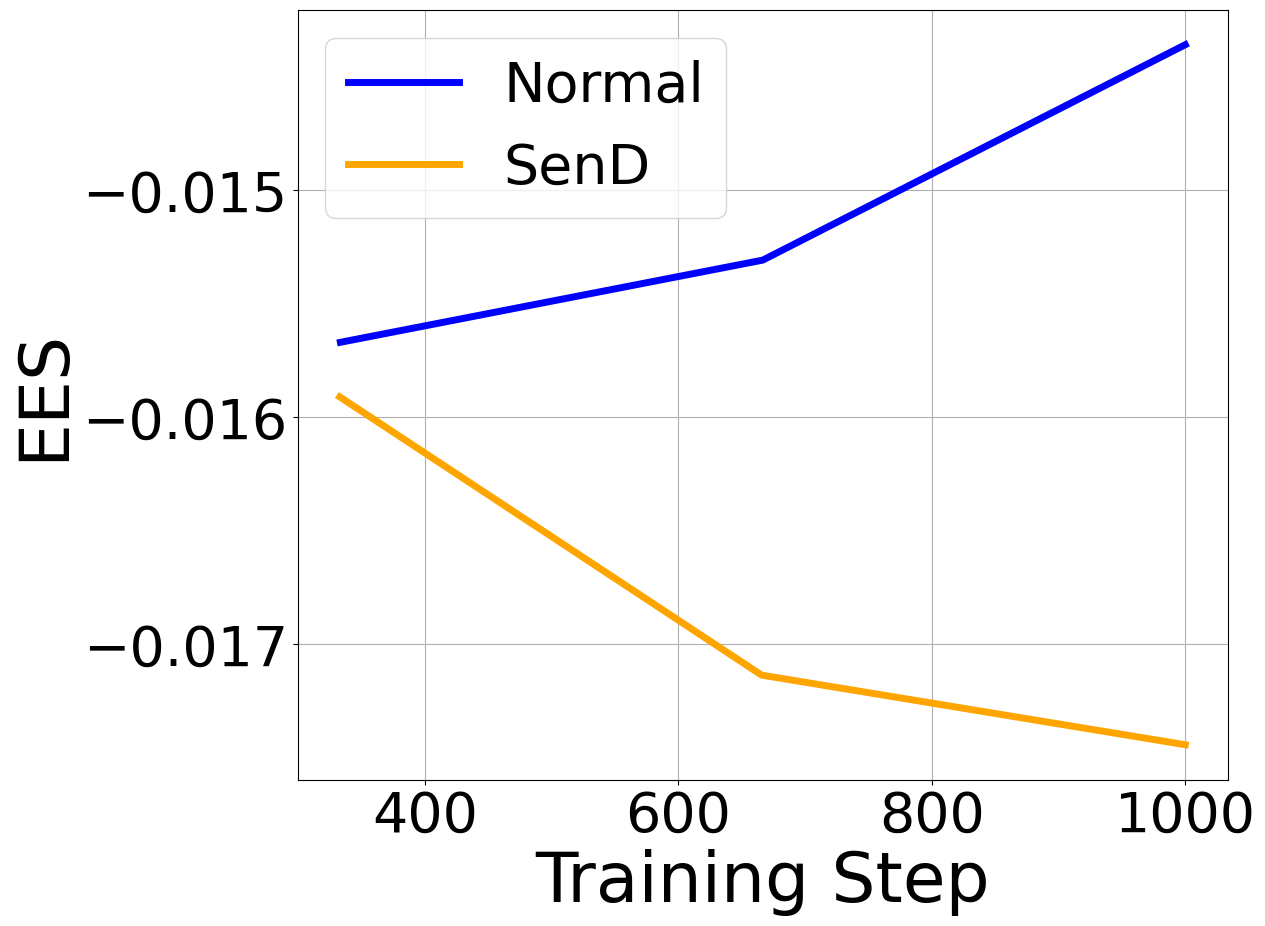}
    \caption{Llama 8B - MedHalt}
    \label{fig:llama_8b_medhalt}
\end{subfigure}
\hfill
\begin{subfigure}[b]{0.25\textwidth}
    \centering
    \includegraphics[width=\textwidth]{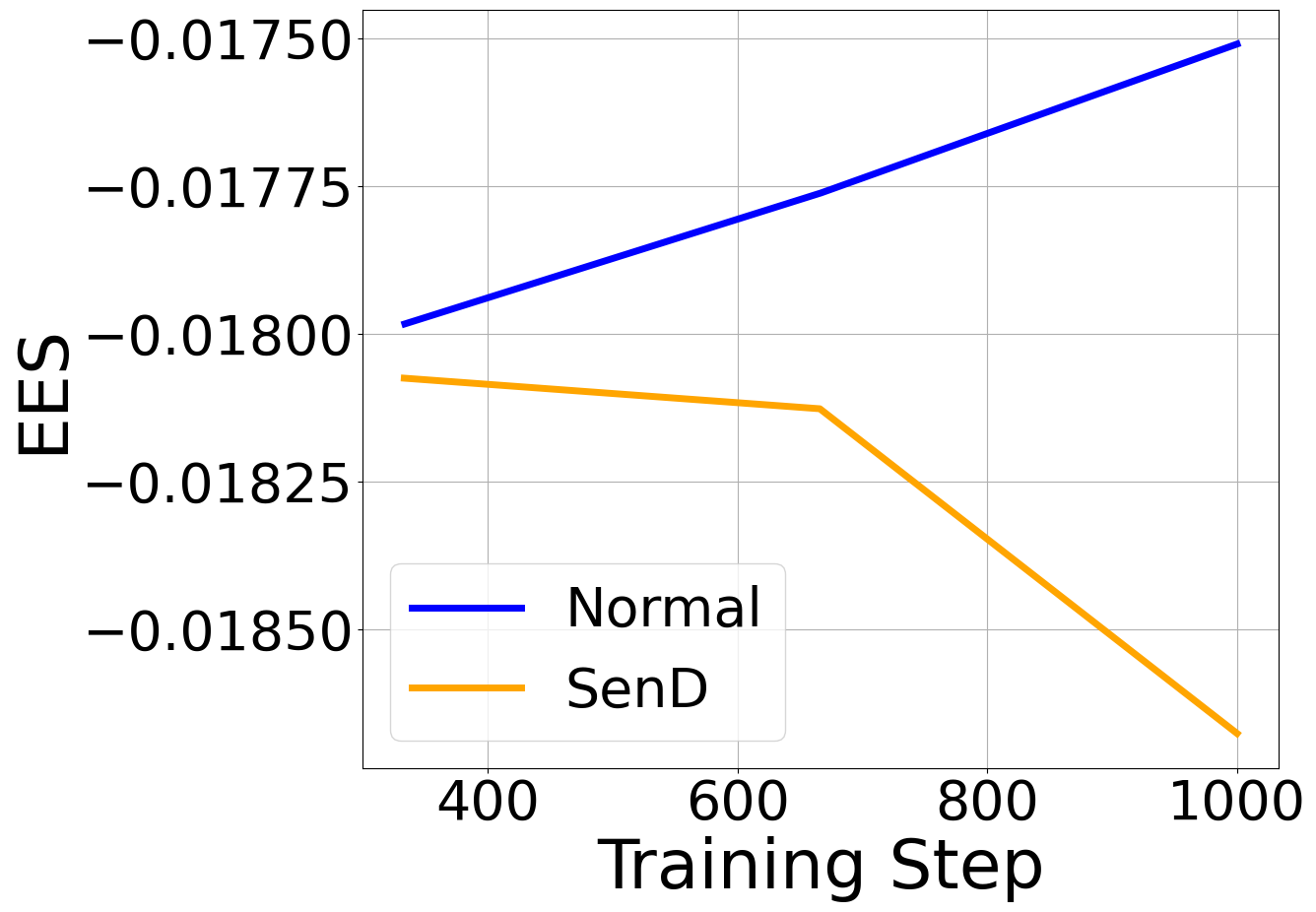}
    \caption{Llama 8B - CodeSearchNet}
    \label{fig:llama_8b_codesearchnet}
\end{subfigure}

\vspace{0.5cm} 

\begin{subfigure}[b]{0.24\textwidth}
\centering
  \includegraphics[width=\textwidth]{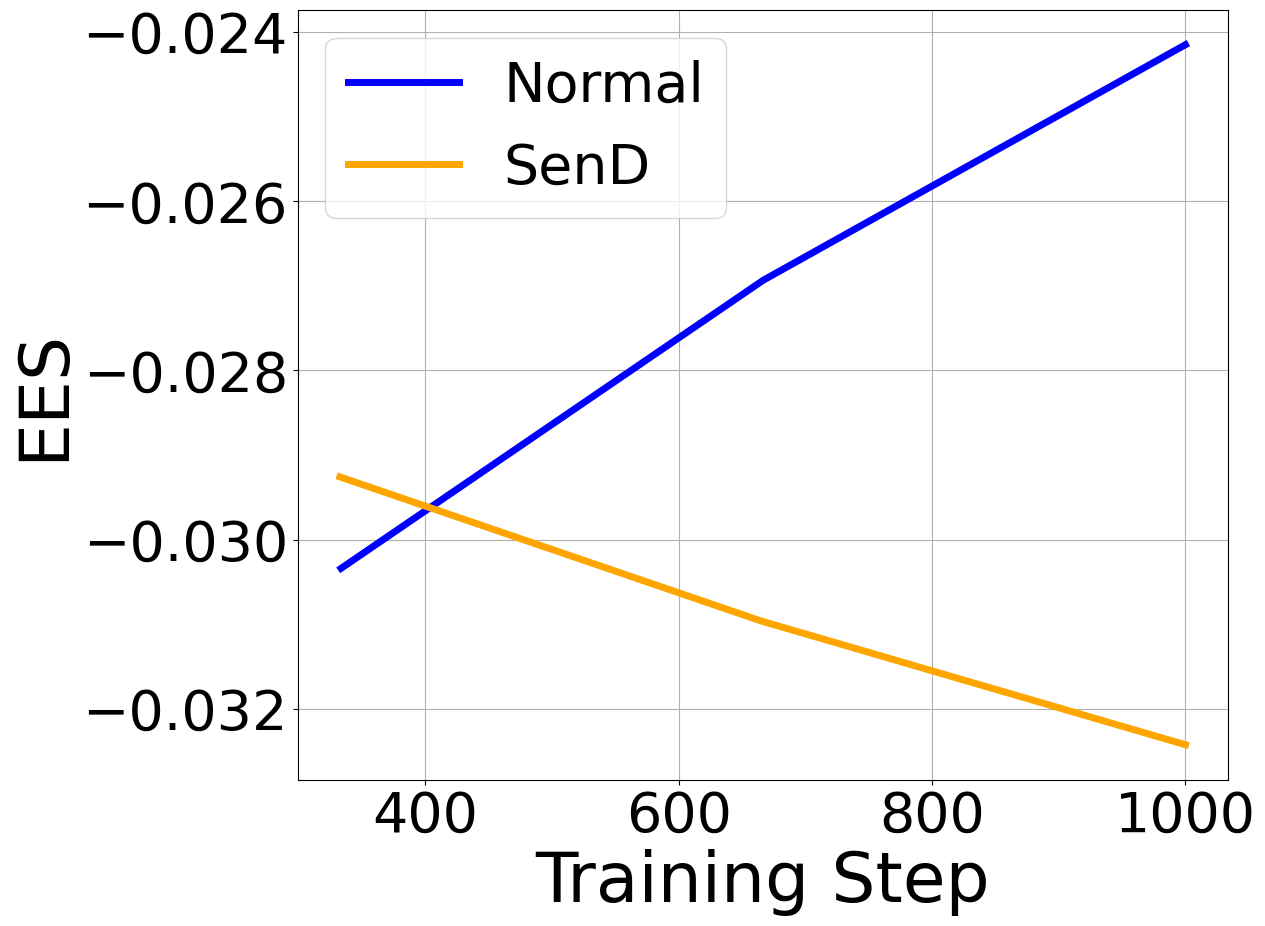}
  \caption{Pythia 1B - HELM}
  \label{fig:pythia_helm}
\end{subfigure}
\hfill
\begin{subfigure}[b]{0.24\textwidth}
\centering
  \includegraphics[width=\textwidth]{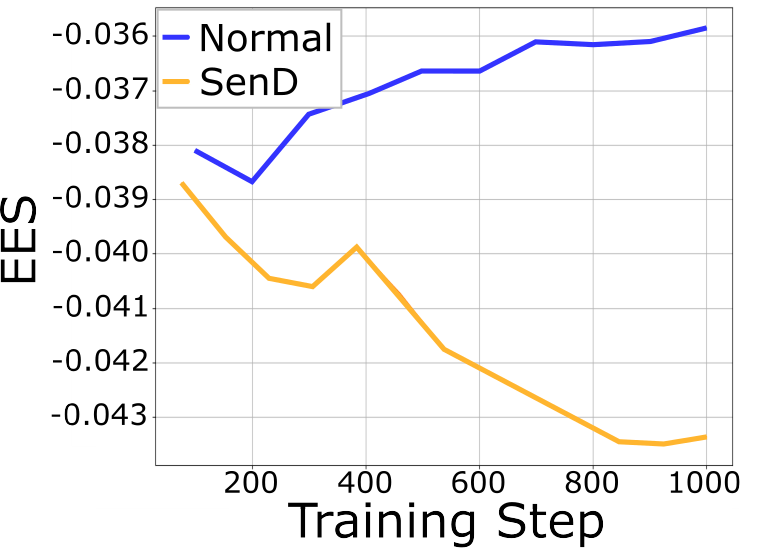}
  \caption{Pythia 1B - LegalBench}
  \label{fig:pythia_legalbench}
\end{subfigure}
\hfill
\begin{subfigure}[b]{0.26\textwidth}
    \centering
    \includegraphics[width=\textwidth]{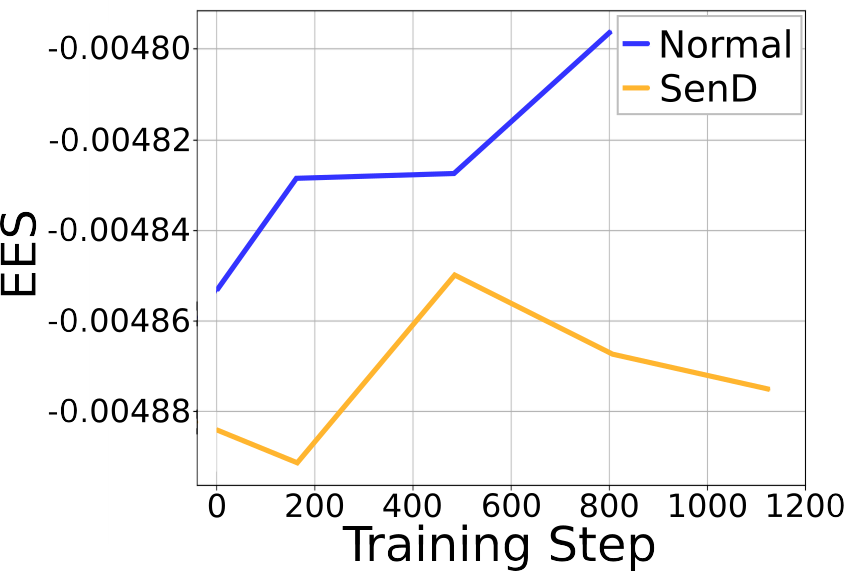}
    \caption{Pythia 1B - MedHalt}
    \label{fig:pythia_1b_medhalt}
\end{subfigure}
\hfill
\begin{subfigure}[b]{0.24\textwidth}
    \centering
    \includegraphics[width=\textwidth]{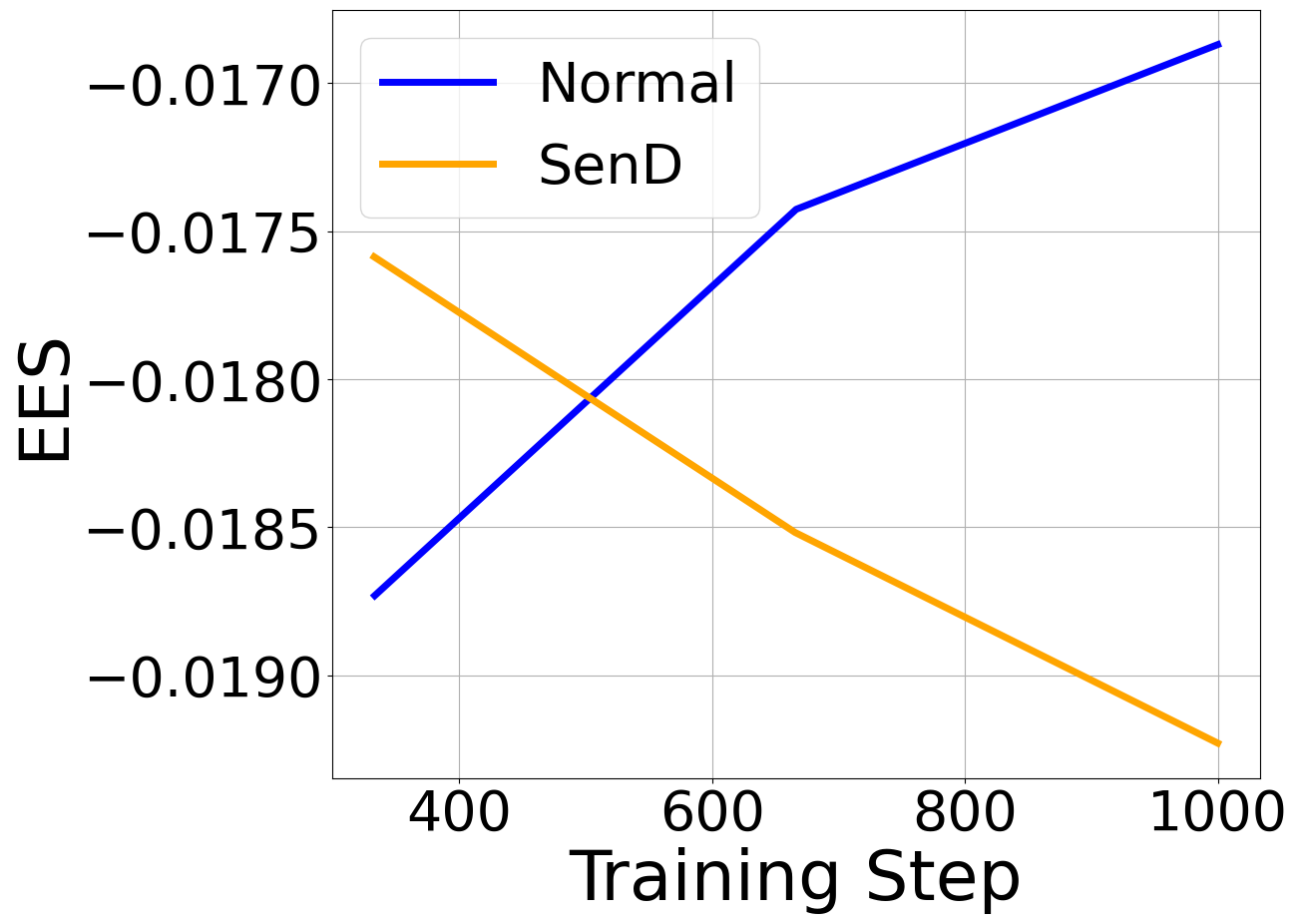}
    \caption{Pythia 1B - CodeSearchNet}
    \label{fig:pythia_1b_codesearchnet}
\end{subfigure}

\caption{\textbf{Regular Training vs. SenD  on HELM and LegalBench datasets.} The first row represents Llama 3.1 8B while the second row shows Pythia 1B models. Column one (a) and (e) is trained on the HELM dataset. Column two (b) and (f) is trained on LegalBench. Column three (c) and (g) use the MedHalt dataset. Column four (d) and (h) are trained on CodeSearchNet. In all cases training with SenD demonstrates a more controlled reduction in \textbf{EES}, optimizing for hallucination mitigation and loss stability. For results on Llama 3.2 1B training, refer to Appendix \ref{appendix:send_more_experiments}.}
\label{fig:finetuning}
\end{figure*}

\section{Sensitivity Dropout (SenD)}
\label{sec:send}
Building on the findings from Section \ref{sec:sensitive_impact}, and aiming to reduce hallucination variance during LLM training, this section introduces SenD, an efficient framework for training LLMs. SenD integrates the EES method discussed in Section \ref{sec:ees} to enhance computational efficiency while addressing variance in SEI behavior. By identifying SEIs, which contribute to the oscillatory behavior of hallucinations during training, SenD deterministically drops these indices based on a small subset of the training data. This approach ensures an increase in the model's response certainty by the end of training as explained in Algorithm \ref{alg:send}.

\begin{algorithm}
    \caption{ Sensitivity Dropout }
    \label{alg:send}
    \small
    \begin{algorithmic}[1]
        \Require $\epsilon$ denotes the acceptable range for loss convergence and $\delta$ denotes acceptable range for confabulation (EES) convergence
        \State Initialize dataset with $\alpha$\% training $Y_t$ and $(100-\alpha)$\% tracking $Y_s$
        \While {Loss $> \epsilon$ and EES $> \delta$} \Comment{Refer to Algorithm \ref{alg:ees} for EES}
            \For{\textit{t} in T} \Comment{T denotes the number of  checkpoints per SEI calculation}
                \State Train LLM for one  checkpoint  over $Y_t$
                \State Record penultimate layer representations $R_t$ of LLM over $Y_s$
            \EndFor

            \For {\(t \text{ in } T - 1\)}
                \State Compute variability $V_t$ from $R_t$ to $R_{t+1}$ \Comment{Refer to Equation \ref{equation:3}}
            \EndFor
            \State Take average Variability \(V_{avg} = \frac{1}{N_s}\sum_{i=0}^{N_s}V_i\)
            \State $s = K$ most sensitive embedding indices  $\in V_{avg}$ \Comment{Refer to Equation \ref{equation:4}}
            \State Drop  embedding indices  $s$ for next T  checkpoints 
        \EndWhile
    \end{algorithmic}
\end{algorithm}
\normalsize

\subsection{SenD Complexity Analysis}
SenD's additional computational complexity compared to traditional transformer training time complexity per epoch comes from three independent steps, mainly from the attention mechanism and multiple inferences during training:
\begin{enumerate}
    \item Generating penultimate layer activations: $O(c(mN)^2dLT)$ This comes from performing $c$ forward passes through the transformer model, each with complexity $O((mN)^2dL)$ due to the quadratic attention mechanism, repeated $T$ times per epoch where $m,N,d,L$ are the context size, dataset size, attention head dimension, and number of layers respectively.
    \item Computing sensitive embedding indices: $O(IT((mN)c+logI))$ Derived from computing sensitivity for each embedding index $I$ with cost $O((mN)c)$, plus selecting top K\% indices with $O(IlogI)$ sorting cost, repeated $T$ times.
    \item EES stopping criterion: $O(N^2)$ Classical EigenScore computes a full eigen-decomposition with $O(N^3)$ complexity for dense matrices where N is the hidden size of the language model. EES reduces this by using Chebyshev polynomial moments and stochastic trace estimation (Further details available in Appendix \ref{sec:ees_derivation}). By replacing eigen-decomposition with iterative matrix-vector multiplications (each costing $O(N^2)$) and using a fixed number of moments and trace samples, EES achieves $O(N^2)$ time complexity.
\end{enumerate}

After removing insignificant terms, the total additional computational complexity per epoch is $O((mN)^2dLT)$ ultimately giving $O((mN)^2)$. Please note that this is equal to adding multiple validation steps to an epoch of an LLM’s training procedure, implying that no excessive inefficient complexity is introduced by SenD, making it an efficient mitigation technique.

Empirical evaluation of the additional computational complexity of SenD is conducted on the HELM dataset \citep{su_unsupervised_2024} with 2,000 datapoints using the Llama 8B model \citep{dubey_llama_2024}. We observe that for one epoch, Send training takes 61 minutes while normal training takes 55 minutes. However, in the context of adaptation and reducing the risks of hallucination, we believe this 11\% increase is a worthwhile investment.

\begin{table*}[t]
\centering
\begin{tabular}{c|*{4}{cc}}
\toprule
\textbf{Metric} & \multicolumn{2}{c}{\textbf{MedHalt}} & \multicolumn{2}{c}{\textbf{HELM}} & \multicolumn{2}{c}{\textbf{LegalBench}} & \multicolumn{2}{c}{\textbf{CodeSearchNet}} \\
\cmidrule(lr){2-3} \cmidrule(lr){4-5} \cmidrule(lr){6-7} \cmidrule(lr){8-9}
 & SenD & Normal & SenD & Normal & SenD & Normal & SenD & Normal \\
\midrule
\textbf{HellaSwag} & 0.73 & \textbf{0.75} & 0.73 & \textbf{0.74} & \textbf{0.73} & 0.72 & \textbf{0.69} & 0.40 \\
\textbf{MMLU}      & 0.42 & \textbf{0.64} & \textbf{0.67} & 0.65 & 0.56 & \textbf{0.59} & \textbf{0.26} & 0.25 \\
\textbf{Token Entropy} & \textbf{0.32} & 0.33 & \textbf{0.79} & 0.95 & 0.49 & 0.49 & \textbf{0.21} & 0.33 \\
\bottomrule
\end{tabular}
\caption{Effects of training Llama 3.1 8B model on downstream tasks with and without SenD. In HellaSwag and MMLU, a higher score depicts better performance and lower Token Entropy shows higher model confidence.}
\label{table:results}
\end{table*}
\subsection{SenD Experiments}

To evaluate SenD, we use Pythia 1B model \citep{biderman_pythia_2023},  Llama 3.2 1B, and Llama 3.1 8B \citep{dubey_llama_2024} continuing their training on specific datasets rather than restarting pretraining for efficiency. We continually train the models on the following datasets: HELM, consisting of Wikipedia text \citep{su_unsupervised_2024}, MedHALT, a medical dataset emulating real-world entrance exam questions \citep{pal_med-halt_2023}, LegalBench consisting of data for reasoning in LLMs \citep{guha_legalbench_2023}, and CodeSearchNet consisting of programming prompts \citep{husain_codesearchnet_2020}. Note that HELM and MedHALT are specifically designed for hallucination detection/mitigation in LLMs. SenD implements the EigenScore reduction technique from Section \ref{sec:sensitive_impact} and detects SEIs using a 3- checkpoint  window on a specialized hallucination tracking dataset. The distance between checkpoints and the dropout rate $K$ are tunable hyperparameters. Given our ablation study in Appendix \ref{appendix:k_ablation}, we opt for $K=20\%$ and Threshold = 3 for the experiments.  SEIs in the penultimate layer are identified based on their variability across  checkpoints  and are deterministically dropped for the subsequent 3 training  checkpoints. This is repeated at each 3- checkpoint  interval until loss convergence, effectively mitigating hallucination tendencies and oscillations. Since we use SenD in a continual manner, we freeze 24 layers for Llama 8B and 12 layers for Llama 1B and Pythia 1B to reduce the effects of forgetting on both SenD and normal training.

\paragraph{RQ4: How does the performance of SenD compare across Pythia and LLaMA models?}

Pythia and Llama training results are illustrated in Figure \ref{fig:finetuning}. To validate that EES accurately approximates the EigenScore metric; we compare the model's progress during training detailed in Appendix \ref{appendix:ees_comparison}. Upon confirming that, we proceed to compare the performance of Pythia 1B, Llama 3.2 1B, and Llama 3.1 8B trained using normal training to SenD.  As shown in Figure \ref{fig:finetuning} for Llama 8B and Pythia 1B and detailed in Appendix \ref{appendix:send_more_experiments} for Llama 1B, across all three models and domains, training with SenD reduces EES as well as variance during training. In all cases, the final model trained with SenD achieves a lower EES compared to standard training, demonstrating its effectiveness.

\begin{table}[t]
\centering
\begin{tabular}{ccc}
\hline
\textbf{Training}                                              & \textbf{SenD} & \textbf{Normal} \\ \hline
FactScore                                                      & \textbf{0.44} & 0.39          \\ \hline
\begin{tabular}[c]{@{}c@{}}FactScore + RAG\end{tabular}         & \textbf{0.50} & 0.40          \\ \hline
\begin{tabular}[c]{@{}c@{}}HaluEval Accuracy\end{tabular}       & \textbf{0.74} & \textbf{0.74} \\ \hline
\begin{tabular}[c]{@{}c@{}}HaluEval Correctness\end{tabular}    & \textbf{0.98} & \textbf{0.98} \\ \hline
\begin{tabular}[c]{@{}c@{}}HaluEval Exact Match\end{tabular}    & \textbf{0.75} & \textbf{0.75} \\ \hline
\end{tabular}
\caption{Hallucination targeted metrics for Llama 8B. Higher values for all metrics are better. For results in Llama 1B and Pythia 1B refer to Table~\ref{tab:send_vs_normal_all_hallu2}.}
\label{tab:fact_hallu_8b}
\end{table}

\paragraph{RQ5: What is the effect of SenD on downstream tasks and uncertainty metrics?}

To assess the effectiveness of SenD on SOTA factuality metrics and downstream tasks, we evaluate several benchmarks. First, the HellaSwag \citep{zellers_HellaSwag_2019} and Massive Multitask Language Understanding (MMLU) \citep{hendrycks_measuring_2021} benchmarks, implemented via LMEval Harness \citep{eval-harness}, verify that downstream performance is maintained (Table \ref{table:results} for Llama 8B; Table \ref{tab:send_vs_normal_downstream} for Llama 1B). In addition, token distribution entropy \citep{kossen_semantic_2024}, FactScore \citep{min_factscore_2023}, and HaluEval \citep{li_halueval_2023} (Tables \ref{table:results} and \ref{tab:send_vs_normal_all_hallu2}) are used to assess model certainty and factuality, where lower entropy indicates higher confidence, FactScore measures factual retention, and HaluEval evaluates hallucination tendencies in question answering. FactScore and HaluEval are run solely on the HELM dataset due to computational power restrictions. HELM was selected for these tests due to its similarity to the testing functions, giving a more accurate depiction of a training and testing scenario. 

SenD does not degrade downstream performance and increases the end model's confidence. As shown in Tables \ref{table:results} and \ref{tab:send_vs_normal_downstream}, HellaSwag and MMLU scores remain consistent with or without SenD for Llama 8B, confirming stable language understanding. Moreover, the reduced average token distribution entropy observed in Table \ref{table:results} (Llama 8B) and Table \ref{tab:send_vs_normal_all_hallu2} (Llama 1B and Pythia 1B) indicates up to a 17\% increase in test-time confidence. Additionally, FactScore improves by 11\% with SenD compared to standard training without RAG (Table \ref{tab:fact_hallu_8b}) and by 10\% relative to training with RAG during inference, demonstrating better retention of factual knowledge. Finally, the HaluEval metrics experience no change, with both models achieving very high scores on accuracy, correctness, and exact match in Tables \ref{tab:fact_hallu_8b} and \ref{tab:send_vs_normal_all_hallu2} for Llama 8B and Llama 1B respectively. The consistent performance in metrics associated to hallucinations shows that not only does SenD reduce variance in training, but also provides a more confident model at test time.

\paragraph{RQ6: How does SenD perform in comparison to existing hallucination mitigation approaches?}
Since SenD is the first method to focus on Hallucinations during the training of LLMs, there are no baselines or SOTA methods to compare it to. However, one could treat SenD as a post-hoc method and compare it to Retrieval Augmented Generation (RAG) \citep{lewis_retrieval-augmented_2021}. 
As shown in Table \ref{tab:fact_hallu_8b}, when applying RAG to a SenD-trained Llama 8B model, it achieves a higher FactScore than RAG on a normally trained model. Similarly, Pythia 1B and Llama 1B have performance increases on FactScore with SenD compared to their normal counterpart with and without RAG detailed in Appendix Table \ref{tab:send_vs_normal_all_hallu2}. These results indicate that even though SenD does not outperform post-hoc methods, SenD with RAG enhances the end model's hallucination performance compared to RAG on a normally trained model and should therefore be used conjointly with SOTA methods.

\section{Conclusion}

In this paper, we presented a protocol to refine the current training methods of LLMs based on experiments showing oscillatory behaviour with respect to hallucinations throughout training (Figure \ref{fig:dynamics}). To do this we used the internal states of LLMs, specifically the penultimate layer activations during inference on a specialized dataset. We present an initial method of reducing hallucinations based on the principles of EigenScore metrics introduced by \citet{chen_inside_2024}. We showed empirically that our SEI detection method significantly reduces the EigenScore on inference of LLMs throughout various stages of training (Figure \ref{fig:drop}). Following the success of the SEI method, we moved on to the application of a hallucination reduction method on training of Pythia and Llama models in various domains. We show through training with SenD that we are able to fix the oscillatory behaviour initially seen throughout training and reduce the EES of finetuned models as shown in Figure \ref{fig:finetuning} by modifying the internal mechanics of training with \textbf{ Sensitivity  Dropout}. At test time we achieve a 25\% increase in FactScore performance and improvement of other SOTA hallucination detection metrics, verifying that SenD  provides a substantial improvement to current training protocols both during and after training in Tables \ref{table:results}, \ref{tab:fact_hallu_8b}, \ref{tab:send_vs_normal_downstream}, and \ref{tab:send_vs_normal_all_hallu2}.

\section{Limitations}

Due to computational limitations, SenD has only been applied to continual training in this paper. However, the SenD training framework is applicable to all stages of training. We encourage future work to implement SenD on larger training sets, such as pretraining, to see how SenD performs in these environments. To further advance our work, we plan to scale SenD  to larger datasets and models, as current experiments were limited by compute constraints with larger LLMs. Demonstrating SenD's effectiveness on larger open-source models like Meta’s Llama 3.2 405B \citep{dubey_llama_2024} will provide crucial evidence for organizations developing state-of-the-art LLMs to incorporate SenD into their training protocols, ultimately improving model safety. Given that SenD  targets variance reduction during training, we anticipate even greater gains on larger LLMs, where the higher inherent variance may amplify the regularization effect and lead to more significant improvements.

\section{Acknowledgements}

Funding support for project activities has been partially provided by the Canada CIFAR AI Chair. We also express our gratitude to Compute Canada and Mila clusters for their support in providing facilities for our evaluations. We would also like to thank Prakhar Ganesh for his role in developing this work at its early stages. Finally, we would like to thank all our reviewers for their insightful feedback that helped improve the clarity and rigour of our paper.

\newpage

\bibliography{references}

\newpage

\appendix

\section{Additional Experiments}

\subsection{Drastic Embedding Changes leading to Sensitive Embedding Indices}

\begin{figure}[ht]
\begin{center}
  \includegraphics[width=\columnwidth]{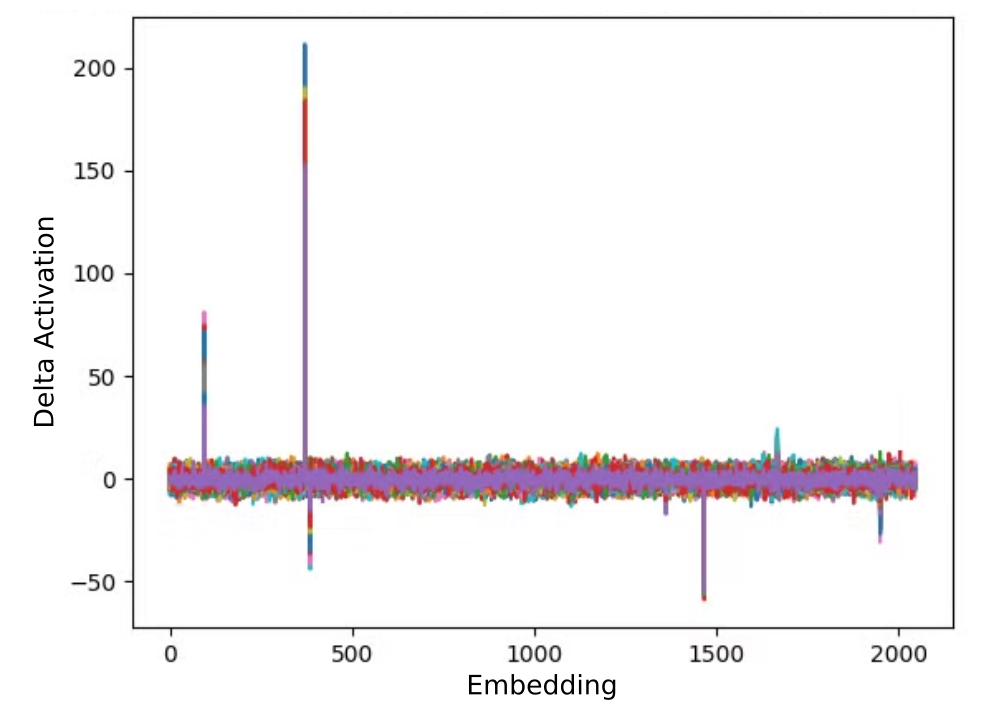}
\end{center}
\caption{\textbf{Net change of sentence embeddings} between checkpoints 125,000 and 143,000. Each different colour is a different input text. As depicted, there are specific  embedding indices  that go through drastic changes between the two checkpoints of the training regardless of the input.}
\label{fig:net_change}
\end{figure}

Looking at internal states of the model allows getting a deeper understanding of the dynamics that could be leading to the oscillatory behaviour seen in Figure \ref{fig:dynamics}. To do this, we record the net change (Definition \ref{definition:net_change_formula}) between checkpoints of the penultimate layer where one checkpoint would be the correct answer and the next would hallucinate. This net change with respect to various different input texts is plotted in Figure \ref{fig:net_change}. It can be observed that there were specific embedding activations that experienced drastically more change relative to the rest of the embeddings. This is the main source of motivation to further define SEIs  (Definition \ref{definition:sensitive_neurons}).
\begin{figure}[ht]
    \centering
    \includegraphics[width=\columnwidth]{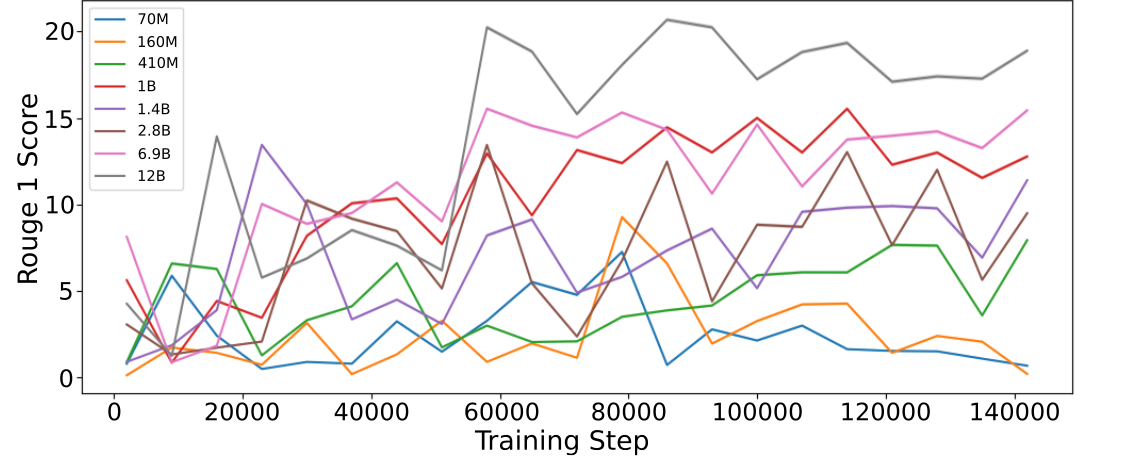} 
    \caption{XSum Rouge 1 Score metric results on Pythia suite.}
    \label{fig:xsum_ablation}
\end{figure}
\begin{figure}[ht]
    \centering
    \includegraphics[width=\columnwidth]{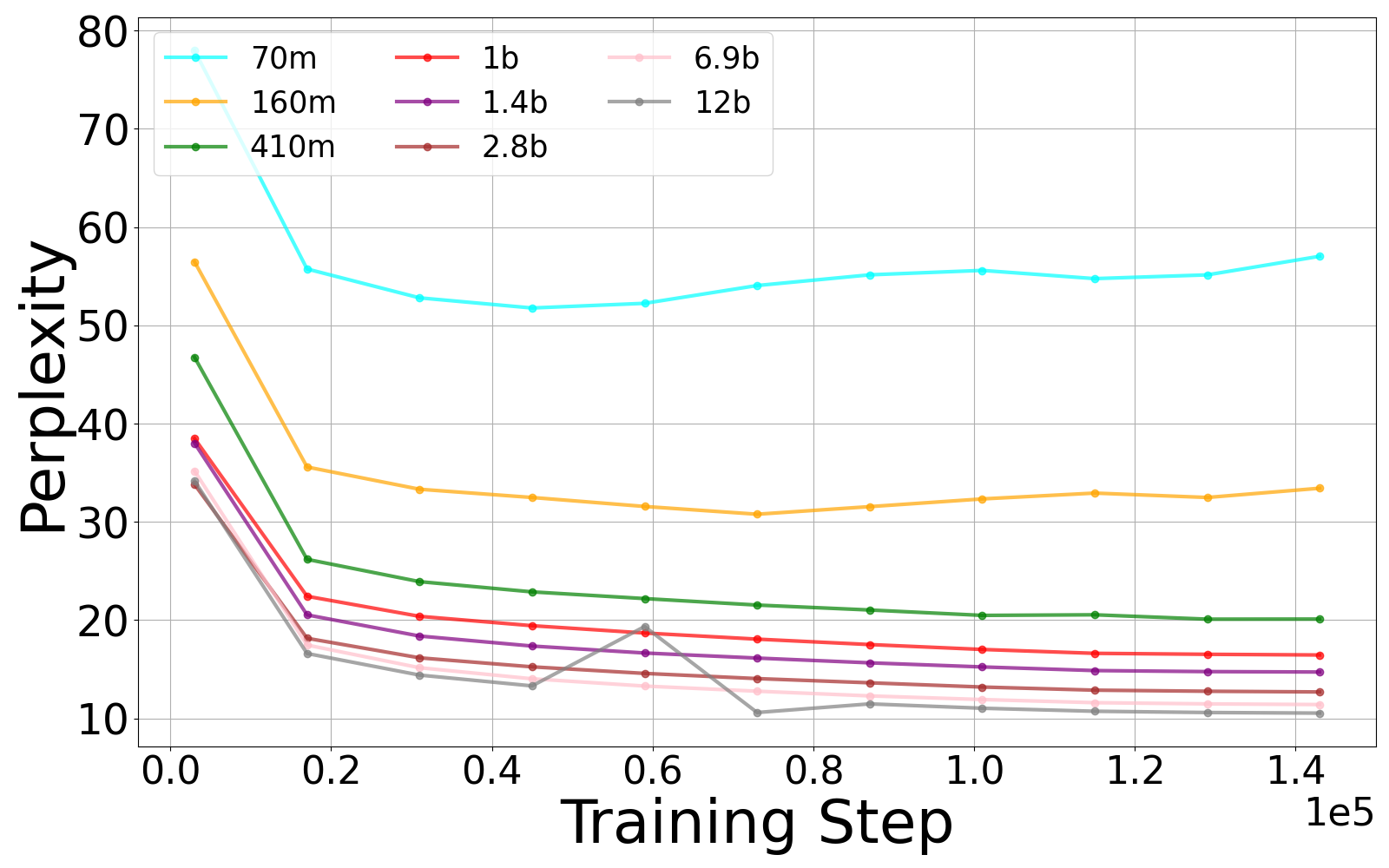} 
    \caption{Perplexity (PPL) metric results on Pythia suite.}
    \label{fig:perplexity_ablation}
\end{figure}

\begin{figure}[ht]
    \centering
    \begin{subfigure}[b]{\columnwidth}
        \centering
        \includegraphics[width=\columnwidth]{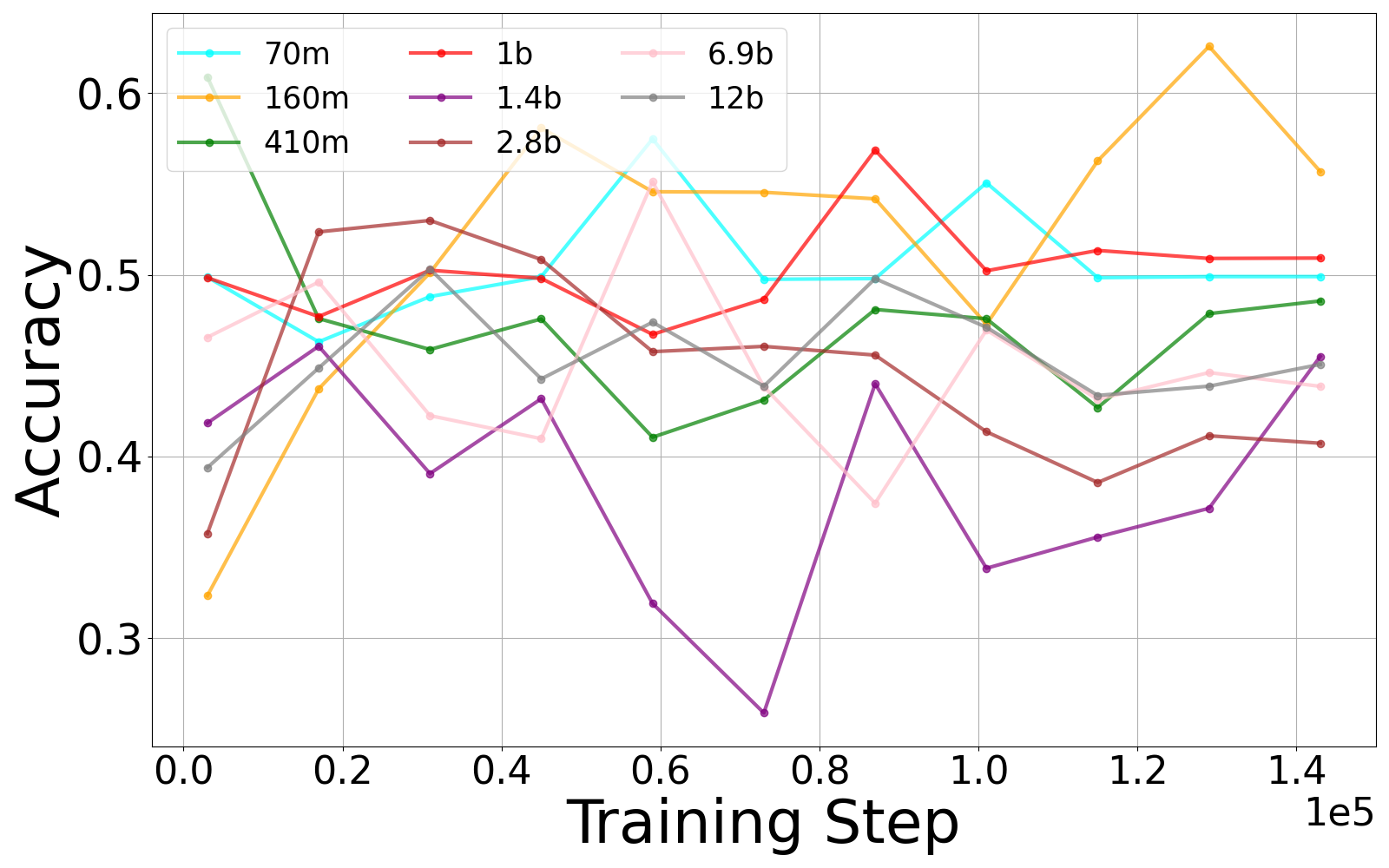} 
        \caption{HaluEval Accuracy metric results.}
        \label{fig:first_subfigure_halu}
    \end{subfigure}
    \hfill
    \begin{subfigure}[b]{\columnwidth}
        \centering
        \includegraphics[width=\columnwidth]{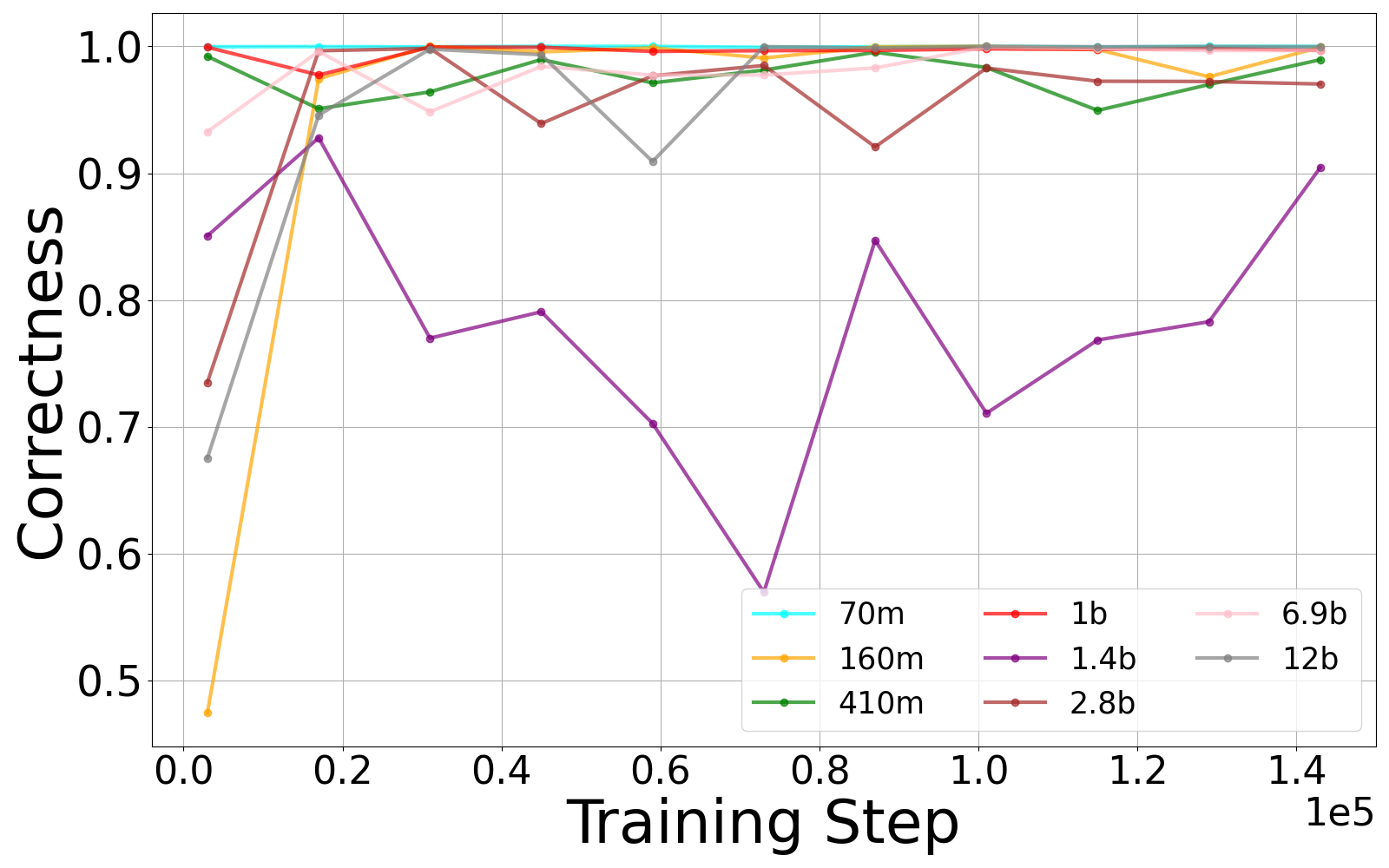} 
        \caption{HaluEval Correctness metric results.}
        \label{fig:second_subfigure_halu}
    \end{subfigure}

    \caption{Ablation studies on various HaluEval metrics for hallucination detection on Pythia suite.}
    \label{fig:halueval_ablation}
\end{figure}

\subsection{Hallucination Oscillations Across Model Sizes}
\label{appendix:oscillations}

Figures \ref{fig:xsum_ablation}, \ref{fig:perplexity_ablation}, and \ref{fig:halueval_ablation} show our study of hallucination oscillations during the training of Pythia models.  An overall observation across the plots is that as opposed to our intuitive expectation which is a linear decrease of the hallucination detection metric when the model scales linearly, neither the oscillations during the training of the model decrease, nor the end model reaches its optimal state in terms of the hallucination metric.

\section{Efficient EigenScore (EES) Derivation}
\label{sec:ees_derivation}
\subsection{Background: Chebyshev polynomials}
\label{appendix:chebyshev}
Chebyshev polynomials are a sequence of orthogonal polynomials in the interval $[-1,1]$ -- orthogonality property shown in equation \ref{equation:cheb_orthogonality} -- that are widely used in numerical analysis, approximation theory, and other areas of applied mathematics. In this work, we are mainly concerned with the Chebyshev polynomials of the first kind with the recurrence relation shown in equation \ref{equation:cheb_recurrence}. Note that this recurrence could also be applied to matrices. Any function $f$ defined in the interval $[-1,1]$ can be approximated with the Chebyshev expansion as shown in \ref{equation:cheb_expansion}.

\begin{multline}
    \label{equation:cheb_orthogonality}
    \int_{-1}^{1} \frac{2}{(1 + \delta_{0n})\pi \sqrt{1 - x^2}} T_m(x) T_n(x) \, dx = \delta_{mn}, \\
    \text{where} \quad \delta_{mn} = 
    \begin{cases} 
    1 & \text{if } m = n, \\
    0 & \text{if } m \neq n,
    \end{cases}
\end{multline}

\begin{equation}
    \label{equation:cheb_recurrence}
    \begin{aligned}
    T_0(x) & = 1, \\
    T_1(x) & = x, \\
    T_{n+1}(x) & = 2x \cdot T_n(x) - T_{n-1}(x), \quad \text{for } n \geq 1.
    \end{aligned}
\end{equation}

\begin{align}
\label{equation:cheb_expansion}
f(x) & = \sum_{n=0}^{\infty} c_n T_n(x), \\
\text{where } c_n & = \frac{2}{\pi} \int_{-1}^{1} \frac{f(x) T_n(x)}{\sqrt{1 - x^2}} \, dx \text{ for } n > 0, \\
c_0 & = \frac{1}{\pi} \int_{-1}^{1} \frac{f(x)}{\sqrt{1 - x^2}} \, dx.
\end{align}

\subsection{Background: DOS and KPM}
\label{appendix:dos}
Let \( H \) be a symmetric matrix \( H \in \mathbb{R}^{N \times N} \) with an eigendecomposition \( H = Q \Lambda Q^T \), where \( \Lambda = \text{diag}(\lambda_1, \cdots, \lambda_N) \) and \( Q = [q_1, \cdots, q_N] \) is orthogonal. The spectral density induced by \( H \) is the generalized function:

\begin{equation}
\mu(\lambda) = \frac{1}{N} \sum_{i=1}^{N} \delta(\lambda - \lambda_i),
\end{equation}

where \( \delta \) is the Dirac delta function. For any analytic test function \( f \), the integral of \( f \) with respect to \( \mu \) is:

\begin{equation}
\label{equation:f_integral}
\int f(\lambda) \mu(\lambda) \, d\lambda = \text{trace}(f(H)).
\end{equation}

\citet{dong_network_2019} introduced KPM as a numerical technique to approximate DOS. KPM approximates DOS by expanding it in terms of chebyshev polynomials. Requiring the matrix's spectrum to be supported in the interval $[-1,1]$, KPM approximates DOS with the following formula, $\lambda$ being the eigen value of the matrix $H$ and $d_m$ approximated by Stochastic Trace Estimation: 

\begin{align}
\mu^{\approx}(\lambda) & = \sum_{m=1}^{\infty} d_m T^*_m(\lambda), \\
\text{where} \quad d_m & = \frac{1}{N} \text{trace}(T_m(H)), \\
\text{and} \quad d_m & \approx \frac{1}{N} \frac{1}{N_z} \sum_{j=1}^{N_z} \mathbf{z}_j^T T_m(H) \mathbf{z}_j, \\
\text{and} \quad T^*_m(x) & = \frac{2}{(1 + \delta_{0m})\pi \sqrt{1 - x^2}} T_m(x).
\end{align}

In the application for hallucination detection, we can use equation \ref{equation:f_integral} to derive a formula for the EigenScore approximation using the properties of Chebyshev polynomials and DOS.

\subsection{Stochastic Trace Estimation on Embedding Matrix}
\label{section:stochastic}
We are interested in computing the $d_m$ term of DOS relying solely on the embedding matrix $E$ therefore we need to rewrite $d_m$ as follows:

\begin{align}
    d_m = \frac{1}{K} \frac{1}{N_z} \sum_{j=0}^{\infty} z_j^TT_m(E^TE)z_j
\end{align}

where $T_m$ can be computed using the Chebyshev polynomials of matrix $ C = E^TE$.

\begin{align*}
T_0(E^TE) \mathbf{z}_j &= I \mathbf{z}_j = \mathbf{z}_j, \\
T_1(E^TE) \mathbf{z}_j &= E^TE \mathbf{z}_j, \\
T_{m+1}(E^TE) \mathbf{z}_j &= 2E^TE T_m(E^TE) \mathbf{z}_j  - ... \\ &... T_{m-1}(E^TE) \mathbf{z}_j
\end{align*}

Each term can be computed with a matrix-vector multiplication.

\subsection{EES Integral Calculation}
\label{section:integral_calculation}
Given the orthogonality of the Chebyshev polynomials, we can simplify the integral mentioned in proposition \ref{proposition:eigenscore}. To approximate the EigenScore, we will expand $\log(\lambda)$ in terms of Chebyshev polynomials and use their orthogonality to simplify the integral.

\textbf{Expanding and Integrating}

To approximate the integral:

\begin{equation}
\frac{1}{K} \int \log(\lambda) \mu(\lambda) \, d\lambda
\end{equation}

Substitute the Chebyshev Expansion for DOS:

\begin{equation}
\mu(\lambda) \approx \sum_{m=0}^{M} d_m T_m^*(\lambda)
\end{equation}

where:

\begin{equation*}
T_m^*(\lambda) = w(\lambda) T_m(\lambda) = \frac{2}{\pi \sqrt{1 - \lambda^2}(1 + \delta_{0m})} T_m(\lambda)
\end{equation*}

Distribute $\log(\lambda)$ in the integral:
\begin{align}
&\frac{1}{K} \int \log(\lambda) \left( \sum_{m=0}^{M} d_m T_m^*(\lambda) \right) \, d\lambda \\ &= \frac{1}{K} \sum_{m=0}^{M} d_m \int \log(\lambda) T_m^*(\lambda) \, d\lambda
\end{align}

\textbf{Evaluate the Integral Using Orthogonality:}

To simplify the integral,

\begin{equation}
\int \log(\lambda) T_m^*(\lambda) \, d\lambda
\end{equation}

First, express $\log(\lambda)$ as a series of Chebyshev polynomials:

\begin{equation}
\log(\lambda) = \sum_{m=0}^{\infty} c_m T_m(\lambda)
\end{equation}

Then:

\begin{equation}
\begin{aligned}
&\int_{0}^{1} \log(\lambda) T_m^*(\lambda) \, d\lambda \\&= \int_{0}^{1} \left( \sum_{m=0}^{\infty} c_m T_m(\lambda) \right) T_m(\lambda) \, d\lambda
\end{aligned}
\end{equation}

Note: The lower bound of the integral is 0 as the matrix is defined in the spectrum $[0,1]$.

Using the orthogonality, we get:

\begin{equation}
\label{equation:ees_main}
c_m = \int_{0}^{1} \log(\lambda) T_m^*(\lambda) \, d\lambda
\end{equation}

So the integral simplifies to:

\begin{equation}
\label{equation:final_ees}
\frac{1}{K} \sum_{m=0}^{M} d_m c_m
\end{equation}

\begin{figure}[ht]
    \centering
    \includegraphics[width=\columnwidth]{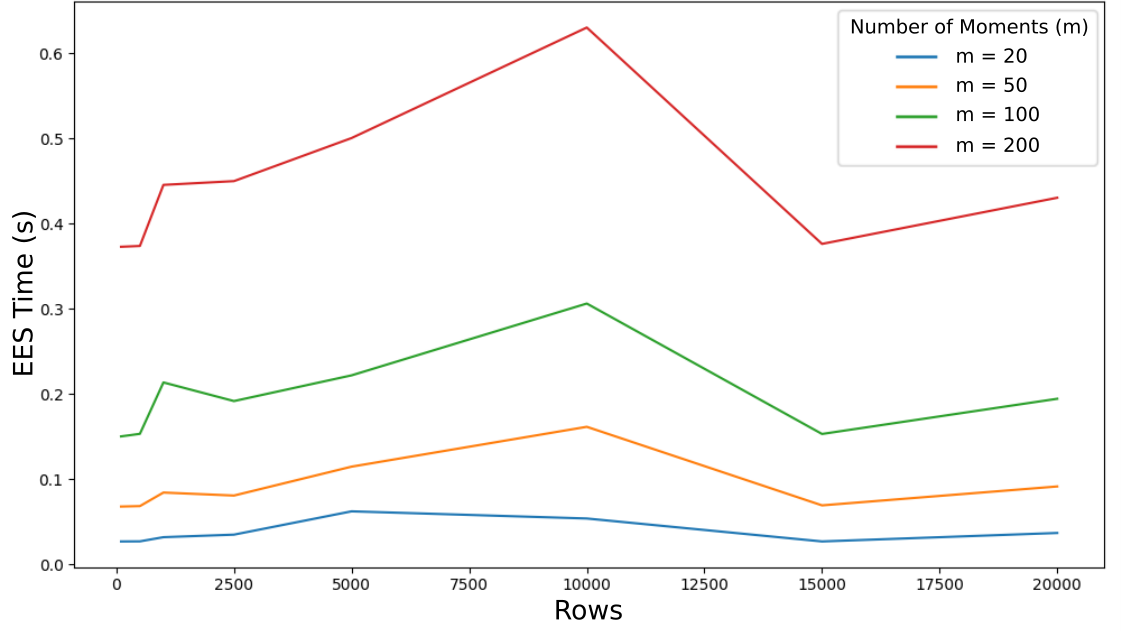}
\caption{\textbf{Effect of changing number of moments on EES} calculation time (seconds). More moments gives more accurate approximation but higher computation time.}
\label{fig:eesMoments}
\end{figure}
\subsection{Efficient EigenScore Moments}

 Figure \ref{fig:eesMoments} presents the effect of using different moment values as the number of matrix rows increases with respect to time. This is an important hyperparameter to tune as increasing the number of moments on EES correlates to having a more accurate and representative approximation of the EigenScore. We observe that as moments in EES increase, the time to calculate EES increases. From this result, we conclude that selecting a moment value of under 50 would provide a balanced trade-off between accuracy and calculation time.

\subsection{EigenScore and EES training trajectories}

\label{appendix:ees_comparison}

\begin{figure}[ht]
    \centering
    \includegraphics[width=\columnwidth]{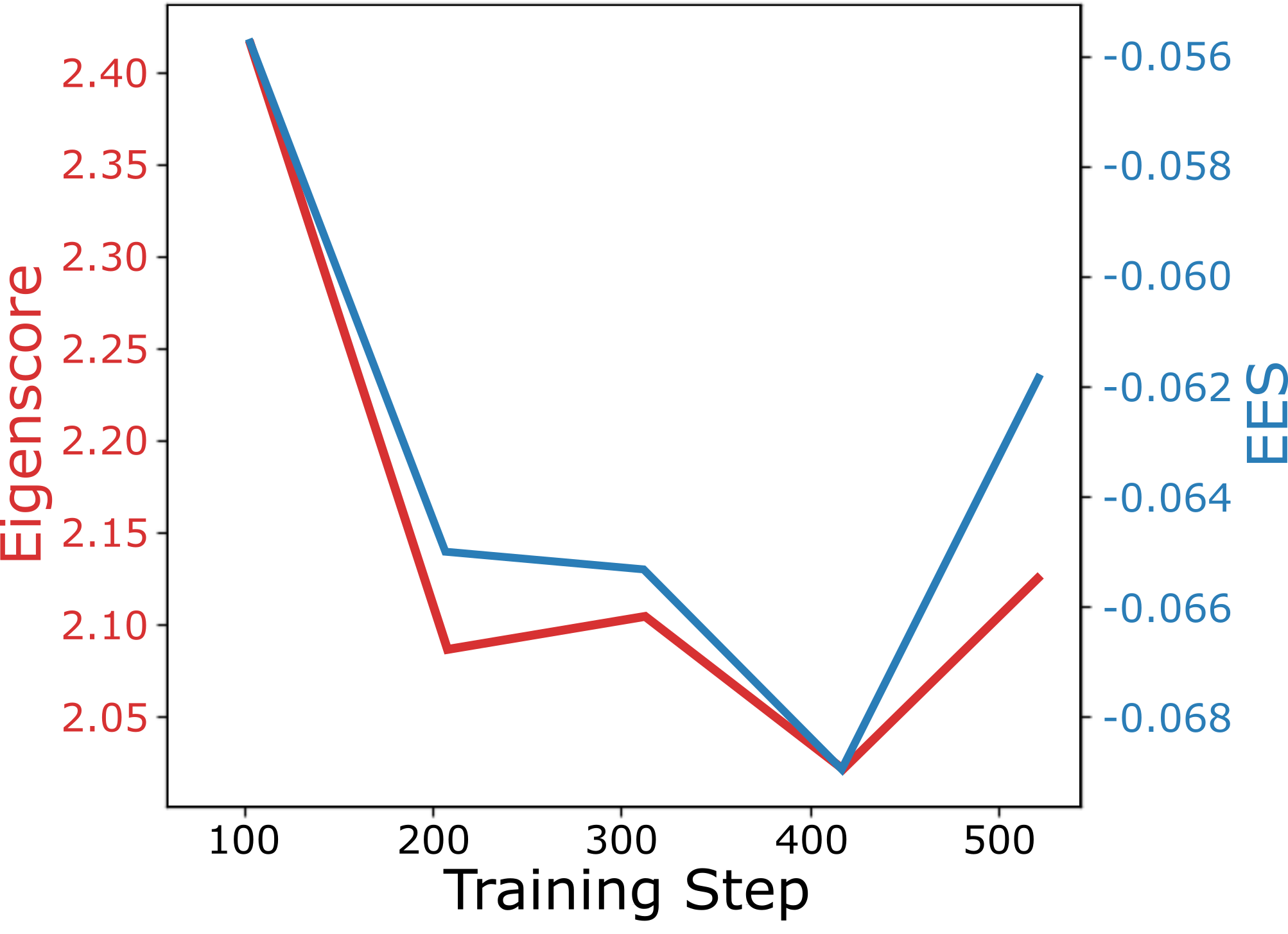}
    \caption{Performance of SenD  on Pythia 1B wih HELM dataset computed with both EES and regular EigenScore. EES is able to closely track the true EigenScore performance metric, showing that it is a good approximator.}
    \label{fig:finetuning_EigenScore}
\end{figure}

To demonstrate the high correlation between EigenScore and EES, we record the progress of Pythia 1B training on the HELM dataset using both EigenScore and EES hallucination metrics (Figure \ref{fig:finetuning_EigenScore}). Albeit a different scale and window, the trajectories, magnitude and shape of the graphs are nearly identical while EES takes only 4 minutes to calculate and EigenScore takes approximately 8, an astounding 2x increase in compute speed. These results show that our metric closely resembles the target metric while greatly reducing the required computational resources.

\section{Ablation study on K and Step Thresholding for SenD}
\label{appendix:k_ablation}

\begin{figure}[ht]
    \centering
    \includegraphics[width=\columnwidth]{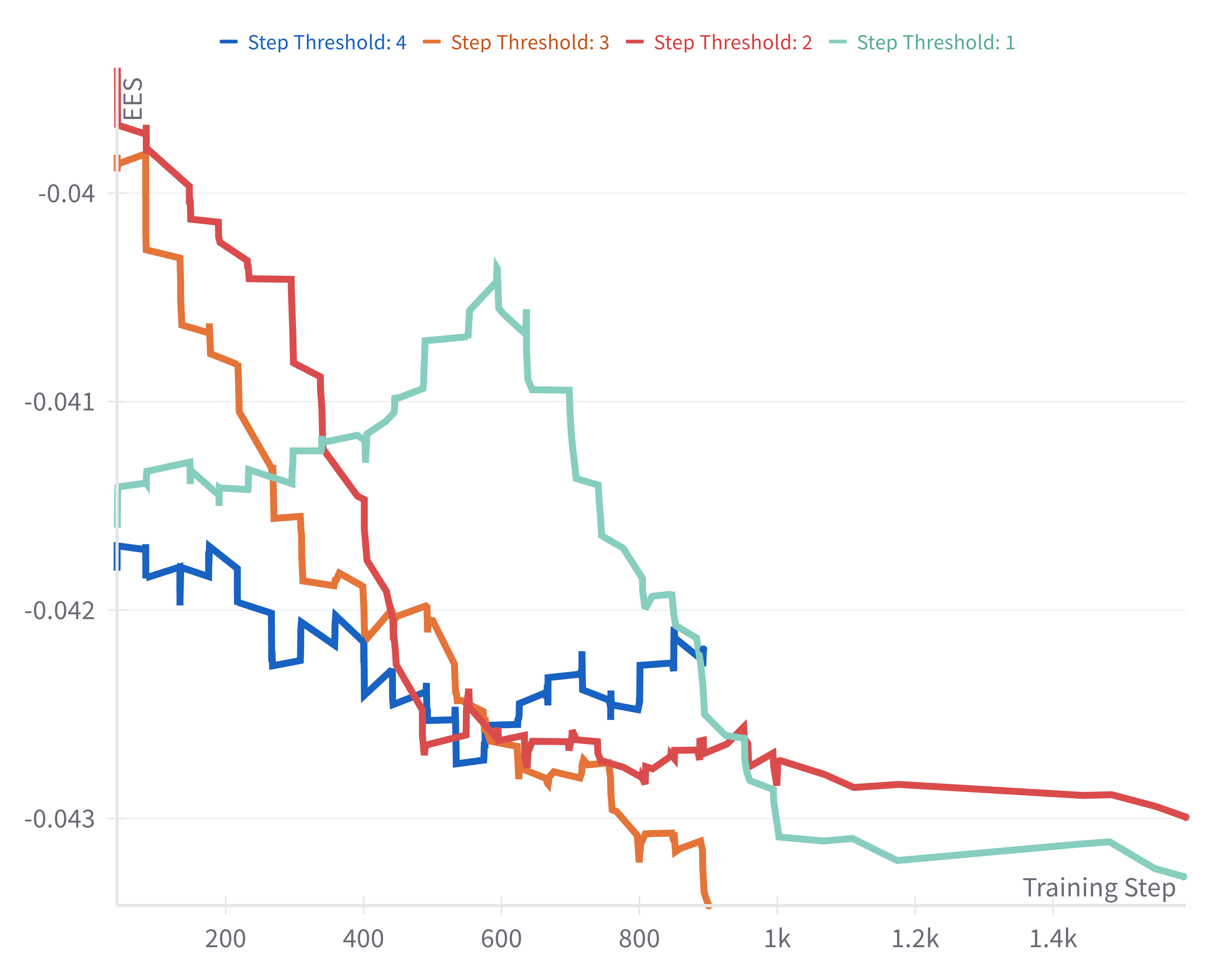} 
    \caption{Ablation on Step Threshold $\in \{1,2,3,4\}$ on the Pythia 1B model with the LegalBench dataset. The fastest drop in EES is achieved by setting Threshold = 3, therefore we choose Threshold = 3 for our experiments. Results are averaged over 5 multiple runs.}
    \label{fig:ablation_step_threshold}
\end{figure}

\begin{figure}[ht]
    \centering
    \begin{subfigure}[b]{\columnwidth}
        \centering
        \includegraphics[width=\textwidth]{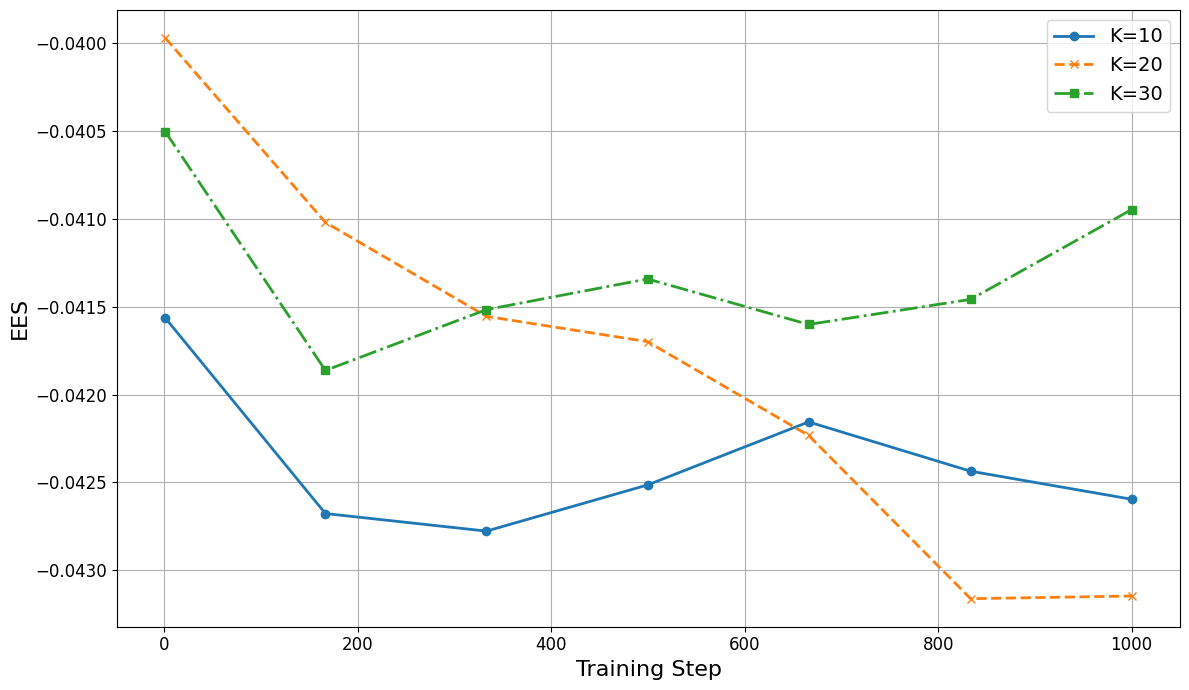} 
        \caption{Ablation study on K using the LegalBench dataset.}
        \label{fig:k_ablation_legal}
    \end{subfigure}
    \hfill
    \begin{subfigure}[b]{\columnwidth}
        \centering
        \includegraphics[width=\textwidth]{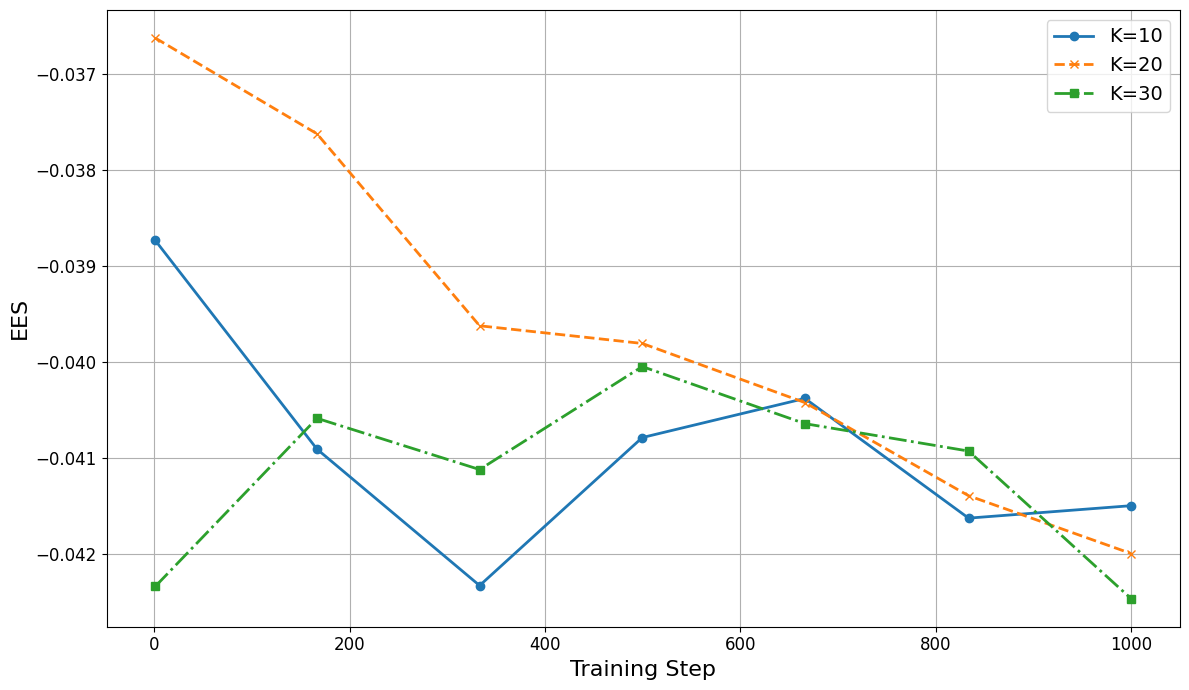} 
        \caption{Ablation study on K using the MedHalt dataset.}
        \label{fig:k_ablation_medhalt}
    \end{subfigure}
    \hfill
    \begin{subfigure}[b]{\columnwidth}
        \centering
        \includegraphics[width=\textwidth]{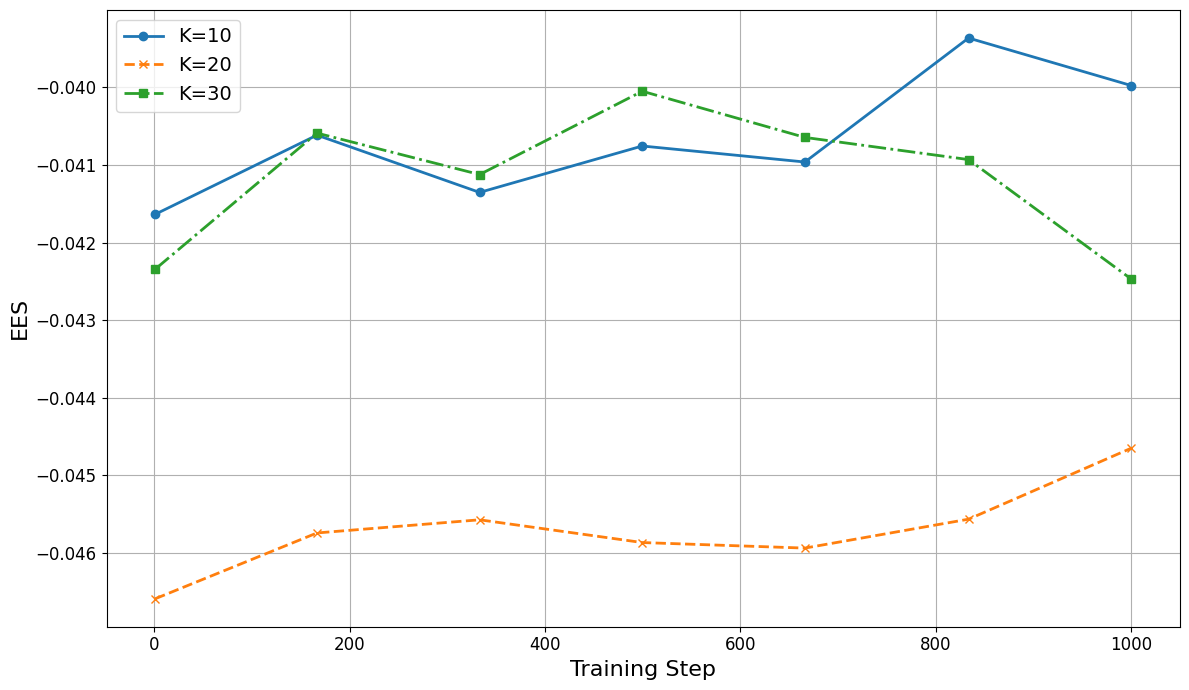} 
        \caption{Ablation study on K using the HELM dataset.}
        \label{fig:k_ablation_helm}
    \end{subfigure}
    
    \caption{Ablation on dropout rate $K \in \{10\%  \text{ orange}, 20\%  \text{ blue}, 30\%  \text{ green}\}$ using the Pythia 1B model averaged over 10 runs on the LegalBench dataset. $K=20\%$ achieves optimal performance in reducing EES throughout training for HELM and LegalBench and almost equalizes $K=30\%$ in stabilizing the halluciantion oscillations, therefore we choose $K=20\%$ for our experiments.}
    \label{fig:ablation_k_combined}
\end{figure}

Figure \ref{fig:ablation_k_combined} shows the ablation study done on $K$ and Figure \ref{fig:ablation_step_threshold} illustrates the ablations study done on the Step Threshold for SenD  experiments. As depicted, $K=20\%$ and Threshold = 3 are chosen for our experiments except for Llama 3.1 8B model which due to its larger size requires more  embedding indices  to be dropped, hence adapting to $K=30\%$.

\section{Additional Pythia 1B, Llama 3.2 1B, and Llama 3.1 8B Training with and without SenD}
\label{appendix:send_more_experiments}

\begin{figure}[t]
\centering
\begin{subfigure}[t]{\linewidth}
\centering
  \includegraphics[width=0.8\linewidth]{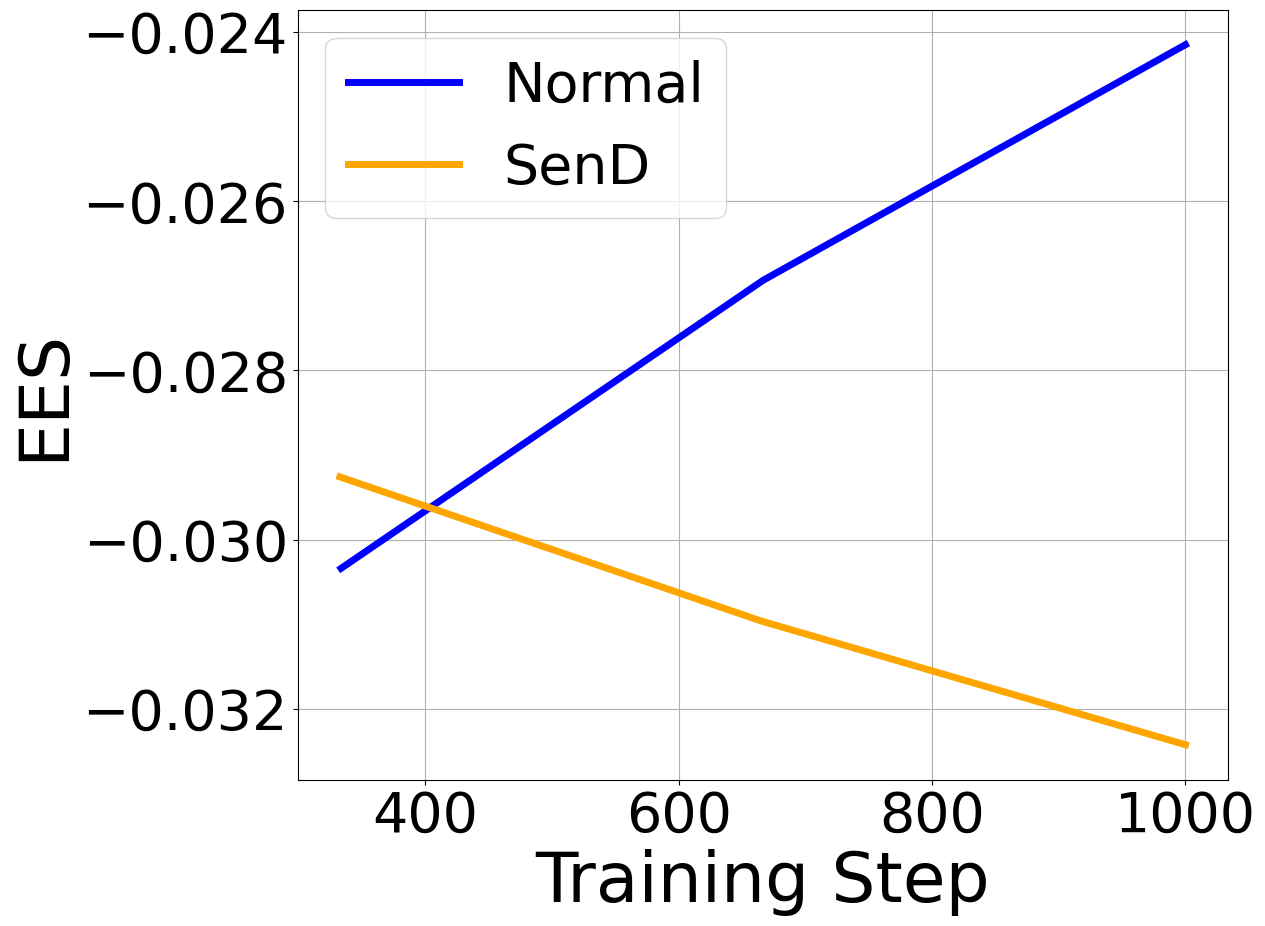}
  \caption{Llama 1B - HELM}
  \label{fig:llama1b_helm}
\end{subfigure}
\hfill
\begin{subfigure}[t]{\linewidth}
\centering
  \includegraphics[width=0.8\linewidth]{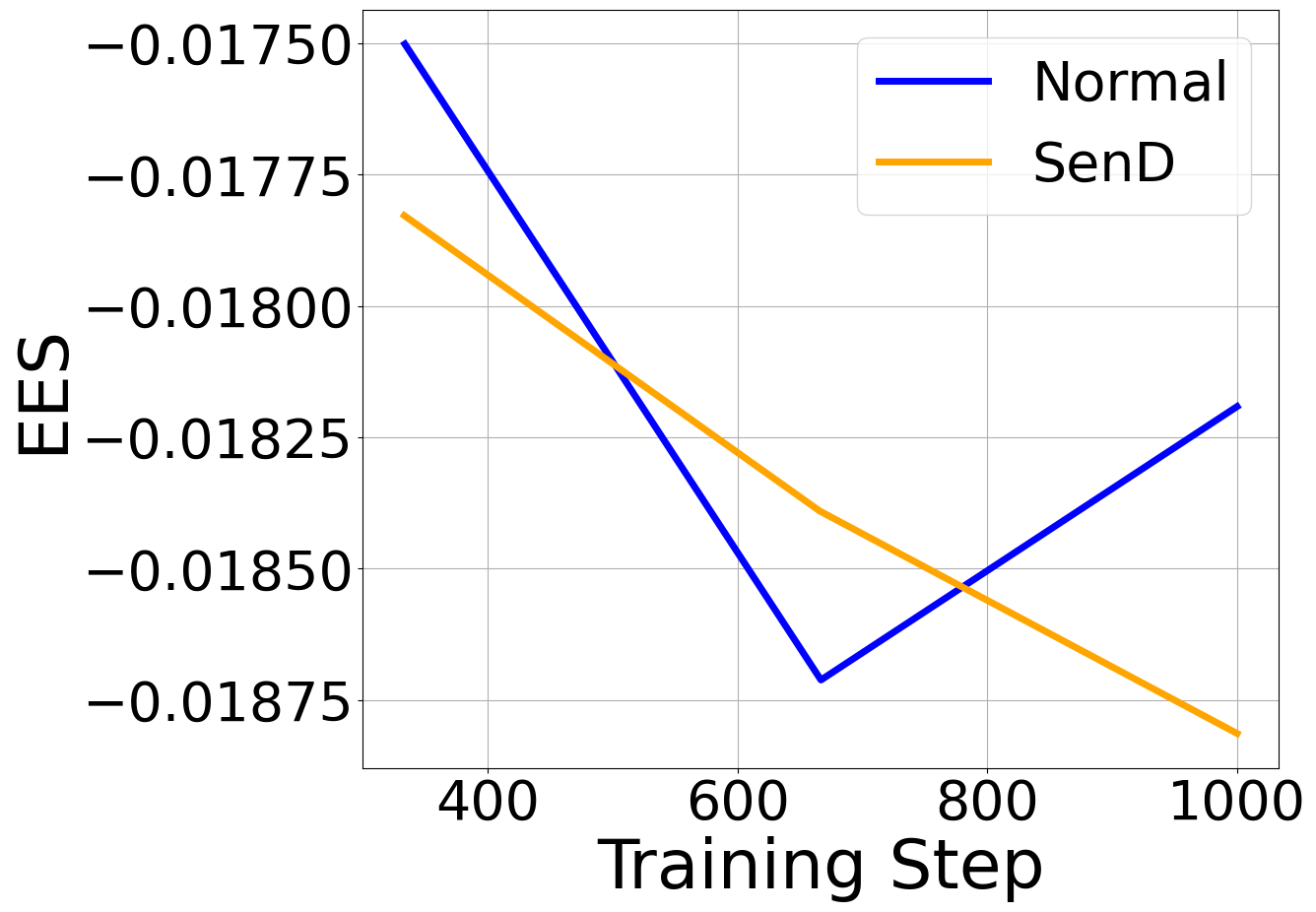}
  \caption{Llama 1B - LegalBench}
  \label{fig:llama1b_legalbench}
\end{subfigure}

\vspace{0.5cm}

\begin{subfigure}[t]{\linewidth}
\centering
  \includegraphics[width=0.8\linewidth]{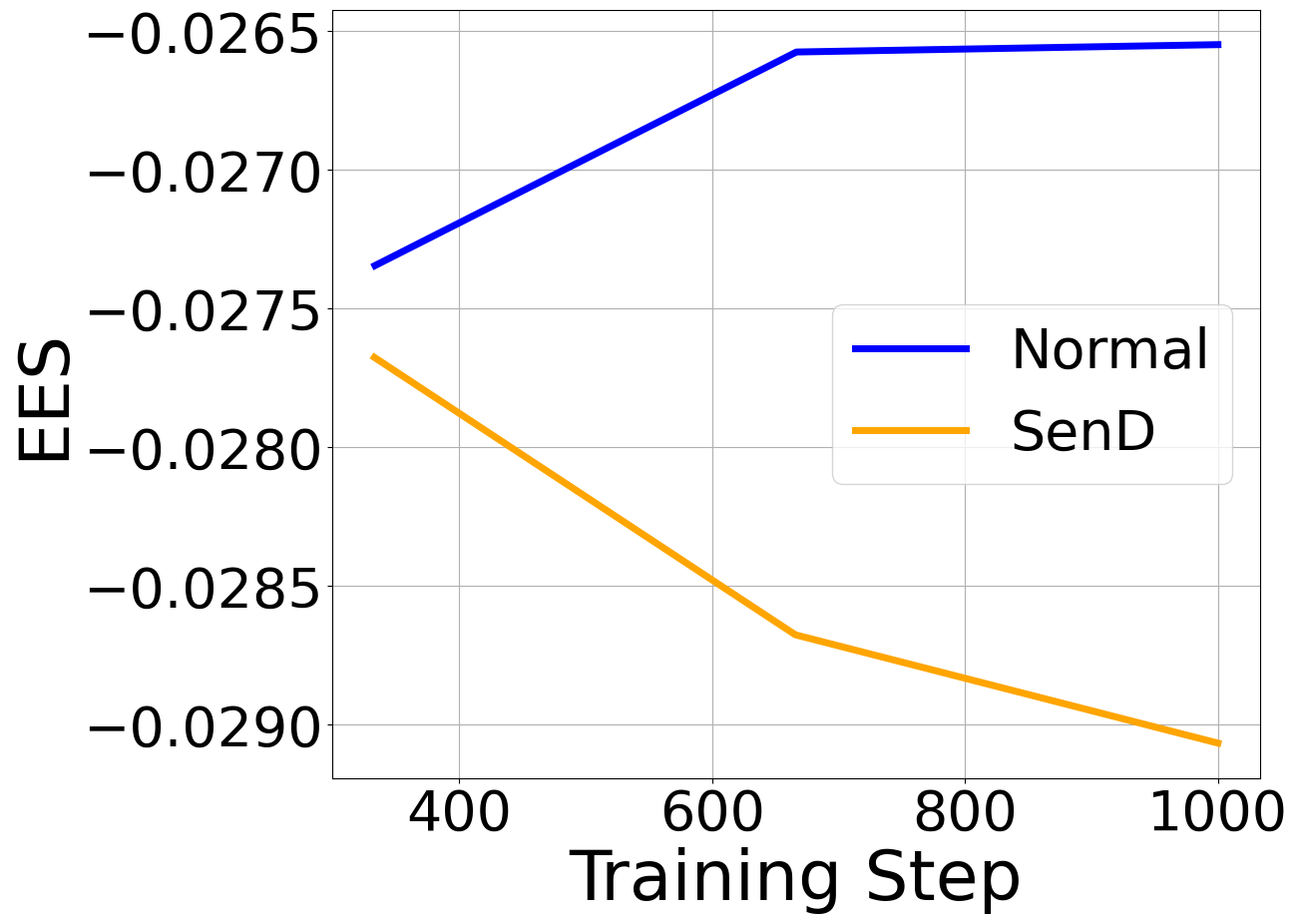}
  \caption{Llama 1B - MedHalt}
  \label{fig:llama1b_medhalt}
\end{subfigure}
\hfill
\begin{subfigure}[t]{\linewidth}
\centering
  \includegraphics[width=0.8\linewidth]{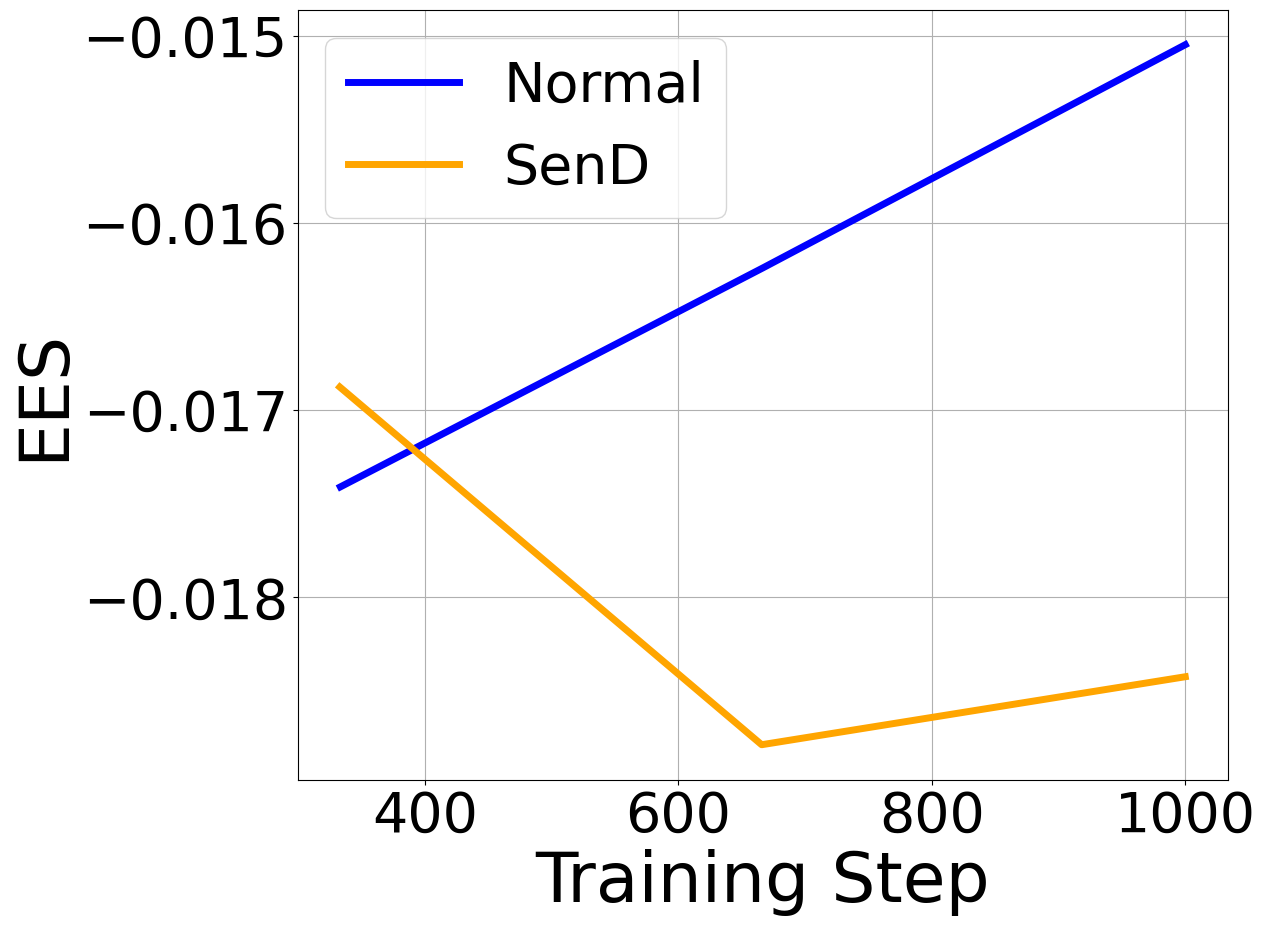}
  \caption{Llama 1B - CodeSearchNet}
  \label{fig:llama1b_codesearchnet}
\end{subfigure}

\caption{Evaluation results for Llama 1B across different benchmarks. SenD consistently outperforms normal training by reducing EES in a more controlled manner.}
\label{fig:llama1b_results}
\end{figure}

Here, we present additional experimental results of training Pythia and Llama on multiple domains. Figure~\ref{fig:llama1b_results} supplements the results discussed in Section~\ref{sec:send} by illustrating the training procedures on an additional model, Llama 1B

In the Pythia 1B setting, the EES achieved with training using SenD remains consistently lower than that of normal training and exhibits fewer oscillations throughout the training process. In the Llama 3.1 8B setting, while both approaches show an increase in the EES metric during training, the final model trained with SenD achieves a lower EES, indicating a reduced likelihood of hallucinations in this domain.

\section{SenD performance across different models, datasets, and metrics}

Here we present a more in depth look at SenD performance, looking at its effect beyond just Llama 3.1 8B. Here we present in Table \ref{tab:send_vs_normal_downstream} the results from running downstream tasks HellaSwag and MMLU are presented for Llama models with sizes 8B and 1B. we can see that although the normally trained models are performing better, the scale of performance increase is negligible, in most cases being within 2\% of SenD's performance. Given this negligible difference, we are confident that the SenD tuning is not drastically affecting the model's performance on downstream tasks. We also present in Table \ref{tab:send_vs_normal_all_hallu2} the results of hallucination based metrics for Llama 8B, Llama 1B, and Pythia 1B. We see that although the HaluEval metrics do not change, they are very good for both models. On the other hand, FactScore is significantly increased when using SenD both with and without RAG. This demonstrates SenD's ability to produce factual information more consistently and the additive power of using both SenD and RAG together.

\begin{table*}
\centering
\begin{tabular}{ccccc}
\hline
\textbf{Model}       & \textbf{Task}                  & \textbf{Training} & \textbf{HellaSwag} & \textbf{MMLU} \\ \hline
\multirow{8}{*}{Llama 8B} & \multirow{2}{*}{MedHalt}       & SenD              & 0.73               & 0.42          \\
                          &                                & Normal              & \textbf{0.75}      & \textbf{0.64} \\ \cline{2-5} 
                          & \multirow{2}{*}{HELM}          & SenD              & 0.73               & \textbf{0.67} \\
                          &                                & Normal              & \textbf{0.74}      & 0.65          \\ \cline{2-5} 
                          & \multirow{2}{*}{LegalBench}    & SenD              & \textbf{0.73}      & 0.56          \\
                          &                                & Normal              & 0.72               & \textbf{0.59} \\ \cline{2-5} 
                          & \multirow{2}{*}{CodeSearchNet} & SenD              & \textbf{0.69}      & \textbf{0.26} \\
                          &                                & Normal              & 0.40               & 0.25          \\ \hline
\multirow{8}{*}{Llama 1B} & \multirow{2}{*}{MedHalt}       & SenD              & 0.59               & 0.40          \\
                          &                                & Normal              & 0.59               & \textbf{0.43} \\ \cline{2-5} 
                          & \multirow{2}{*}{HELM}          & SenD              & 0.59               & 0.43          \\
                          &                                & Normal              & 0.59               & \textbf{0.44} \\ \cline{2-5} 
                          & \multirow{2}{*}{LegalBench}    & SenD              & 0.57               & 0.34          \\
                          &                                & Normal              & 0.57               & \textbf{0.35} \\ \cline{2-5} 
                          & \multirow{2}{*}{CodeSearchNet} & SenD              & 0.58               & 0.42          \\
                          &                                & Normal              & \textbf{0.59}      & 0.42          \\ \hline
\end{tabular}
\caption{\textbf{Final Model Downstream Performance}: SenD vs. Normal Training on Llama 8B and 1B on downstream tasks HellaSwag and MMLU. A higher score is better for both of these metrics.}
\label{tab:send_vs_normal_downstream}
\end{table*}
    
\begin{table*}
\centering
\begin{tabular}{ccccccc}
\hline
\textbf{Model}                                                 & \multicolumn{2}{c}{\textbf{Llama 8B}} & \multicolumn{2}{c}{\textbf{Llama 1B}} & \multicolumn{2}{c}{\textbf{Pythia 1B}} \\ \hline
Training                                                       & SenD                   & Normal         & SenD              & Normal              & SenD                   & Normal          \\ \hline
FactScore                                                      & \textbf{0.44}          & 0.39         & \textbf{0.35}              & 0.30                  & \textbf{0.07}          & 0.05          \\ \hline
\begin{tabular}[c]{@{}c@{}}FactScore\\ + RAG\end{tabular}      & \textbf{0.50}          & 0.40         & 0.40              & 0.40                 & \textbf{0.28}          & 0.25          \\ \hline
\begin{tabular}[c]{@{}c@{}}HaluEval\\ Accuracy\end{tabular}    & 0.74                   & 0.74         & 0.49              & 0.49              & \textbf{0.016}         & 0.014         \\ \hline
\begin{tabular}[c]{@{}c@{}}HaluEval\\ Correctness\end{tabular} & 0.98                   & 0.98         & 0.99              & 0.99              & 0.027                  & 0.027         \\ \hline
\begin{tabular}[c]{@{}c@{}}HaluEval\\ Exact Match\end{tabular} & 0.75                   & 0.75         & 0.49              & 0.49              & \textbf{0.589}         & 0.496         \\ \hline
\begin{tabular}[c]{@{}c@{}}Entropy of\\ Tokens\end{tabular}    & \textbf{0.79}          & 0.95         & 1.01              & 1.01              & \textbf{1.44}          & 1.49         
\end{tabular}
\caption{\textbf{Final Model Hallucination Performance: SenD vs. Normal Training (Pythia 1B, Llama 8b, and Llama 1B)}. HaluEval refers to a QA task. Differing factors between the two FactScore tasks (100 and 1k) refers to the number of testing points.}
\label{tab:send_vs_normal_all_hallu2}
\end{table*}

\end{document}